\documentclass[runningheads]{llncs}
\usepackage{amsmath}
\usepackage{graphicx}
\usepackage{comment}
\usepackage{amssymb}
\usepackage{float}
\usepackage{array}
\usepackage{multirow}
\usepackage[width=122mm,left=12mm,paperwidth=146mm,height=193mm,top=12mm,paperheight=217mm]{geometry}
\usepackage{marvosym}
\usepackage{multibib}
\newcites{Supp}{Supplementary References}
\usepackage{arydshln}
\usepackage{longtable}

\usepackage{hyperref}
\hypersetup{
    colorlinks=true,
    urlcolor=magenta
}

\begin{document}
\pagestyle{headings}
\mainmatter

\title{Guiding Monocular Depth Estimation Using Depth-Attention Volume}
\titlerunning{Guiding Monocular Depth Estimation Using Depth-Attention Volume}

\author{Lam Huynh\inst{1}\orcidID{0000-0002-8311-1288} \and
Phong Nguyen-Ha\inst{1}\orcidID{0000-0002-9678-0886} \and
Jiri Matas\inst{2}\orcidID{0000-0003-0863-4844} \and
Esa Rahtu\inst{3}\orcidID{0000-0001-8767-0864} \and
Janne Heikkil\"a\inst{1}\orcidID{0000-0003-0073-0866} }
\authorrunning{L. Huynh et al.}

\institute{Center for Machine Vision and Signal Analysis, University of Oulu, Finland 
\email{\{lam.huynh,phong.nguyen,janne.heikkila\}@oulu.fi}
\and
Center for Machine Perception, Czech Technical University, Czech Republic 
\email{matas@fel.cvut.cz} \\ \and
Computer Vision Group, Tampere University, Finland \\
\email{esa.rahtu@tuni.fi} }

\maketitle

\begin{abstract}

Recovering the scene depth from a single image is an ill-posed problem that requires additional priors, often referred to as monocular depth cues, to disambiguate different 3D interpretations. In recent works, those priors have been learned in an end-to-end manner from large datasets by using deep neural networks. In this paper, we propose guiding depth estimation to favor planar structures that are ubiquitous especially in indoor environments. This is achieved by incorporating a non-local coplanarity constraint to the network with a novel attention mechanism called depth-attention volume (DAV). Experiments on two popular indoor datasets, namely NYU-Depth-v2 and ScanNet, show that our method achieves state-of-the-art depth estimation results while using only a fraction of the number of parameters needed by the competing methods. Code is available at: \url{https://github.com/HuynhLam/DAV}


\end{abstract}

\section{Introduction}
Depth estimation is a fundamental problem in computer vision due to its wide variety of applications including 3D modeling, augmented reality and autonomous vehicles. Conventionally it has been tackled by using stereo and structure from motion techniques based on multiple view geometry \cite{hartley2003multiple,szeliski2011structure}. In recent years, the advances in deep learning have made monocular depth estimation a compelling alternative %to those conventional methods
\cite{chen2019structure,eigen2015predicting,fu2018deep,hao2018detail,Hu2018RevisitingSI,laina2016deeper,lee2019monocular,liu2018planenet,qi2018geonet,ramamonjisoa2019sharpnet,ren2019deep,Yin2019enforcing,zhang2019pattern}.

In learning-based monocular depth estimation, the basic idea is simply to train a model to predict a depth map for a given input image, and to hope that the model can learn those monocular cues that enable inferring the depth directly from the pixel values. This kind of a brute-force approach requires a huge amount of training data and leads to large network architectures. It has been a common practice to use a deep encoder such as VGG-16 \cite{eigen2015predicting}, ResNet-50 \cite{laina2016deeper,qi2018geonet,ramamonjisoa2019sharpnet}, ResNet-101 \cite{fu2018deep}, ResNext-101 \cite{Yin2019enforcing}, SeNet-154 \cite{chen2019structure,Hu2018RevisitingSI} followed by some upsampling and fusion strategy including 
the up-projection module \cite{laina2016deeper}, multi-scale feature fusion \cite{Hu2018RevisitingSI} or adaptive dense feature fusion \cite{chen2019structure} that all result in bulky networks with a large number of parameters. Because high computational complexity and memory requirements limit the use of these networks in practical applications, also fast monocular depth estimation models such as FastDepth \cite{wofk2019fastdepth} have been proposed, but their speed increase comes with the price of reduced accuracy. Moreover, despite of good results achieved with standard benchmark datasets such as NYU-Depth-v2, it still remains questionable if these networks are able to generalize well to unseen scenes and poses that are not present in the training data.     

Instead of trying to learn all the monocular cues blindly from the data, in this paper, we investigate an approach where the learning is guided by exploiting a simple coplanarity constraint for scene points that are located on the same planar surfaces. Coplanarity is an important constraint especially in indoor environments that are composed of several non-parallel planar surfaces such as walls, floor, ceiling, tables, etc. We introduce a concept of depth-attention volume (DAV) to aggregate spatial information non-locally from those coplanar structures. We use both fronto-parallel and non-fronto-parallel constraints to learn the DAV in an end-to-end manner.

It should be noticed that plane approximations have already been used previously in monocular depth estimation, for example, in PlaneNet \cite{liu2018planenet}, where 3D planes were explicitly segmented and estimated from the images, but in contrast to these works, we embed the coplanarity constraint {\em implicitly} to the model by using the DAV, which is a building block inspired by the non-local neural networks \cite{wang2018non}. Unlike the convolutional operation, it operates non-locally and produces a weighted average of the features across the whole image paying attention on planar structures, and favoring depth values that are originating from those planes. By using the DAV we not only incorporate an efficient and important geometric constraint to the model, but also enable shrinking the size of the network considerably without sacrificing the accuracy. To summarize, our key contributions include:   

\begin{itemize}
    \item A novel attention mechanism called depth-attention volume that captures non-local depth dependencies between coplanar points.
    \item An end-to-end neural network architecture that implicitly learns to recognize planar structures from the scene and use them as priors in monocular depth estimation. 
    \item State-of-the-art depth estimation results on NYU-Depth-v2 and ScanNet datasets with a model that uses considerably less parameters than previous methods achieving similar performance.
\end{itemize}

\begin{figure}[!t]
  \centering
    \includegraphics[width=0.89\textwidth]{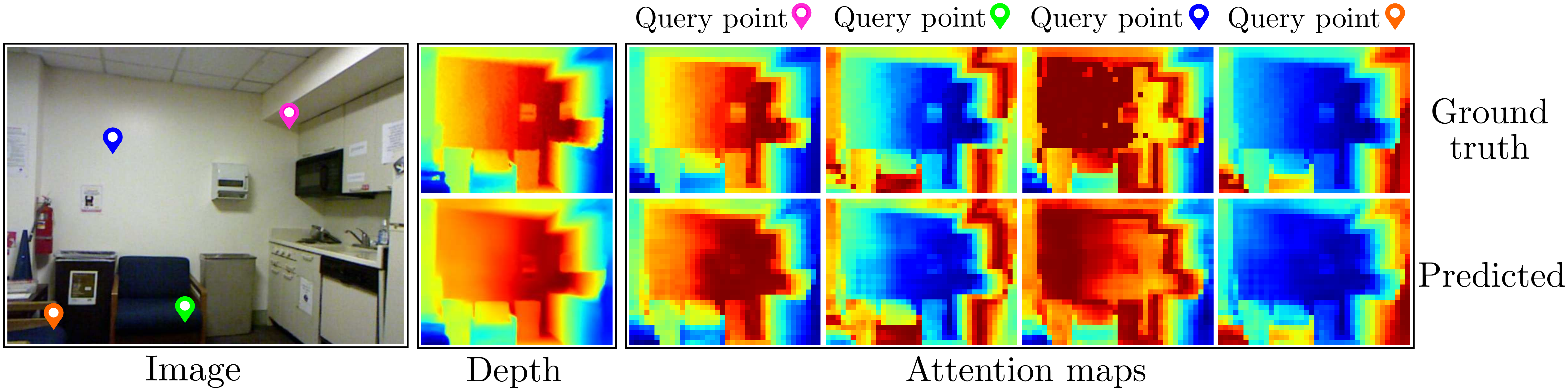}
  \caption{
  Visualization of depth-attention maps. The input image with four query points is shown on the left. The corresponding ground-truth and predicted depth maps are in the middle. Because of the coplanarity prior the depth of the textureless white wall can be accurately recovered. The ground-truth and predicted depth-attention maps for the query points are on the right. Warm colour indicates strong depth prediction ability for the query point.
  }
  \label{fig:dav_illustration}
\end{figure}

\section{Related work}
\textbf{Learning-based monocular depth estimation:} Saxena et al. \cite{saxena2006learning} is one of the first studies using Markov Random Field (MRF) to predict depth from a single image. Later on Eigen et al. proposed method to estimate depth using multi-scale deep network \cite{eigen2014depth} and a multi-task learning model \cite{eigen2015predicting}. Since then, various studies using deep neural networks (DNNs) have been introduced. Laina et al. \cite{laina2016deeper} employed a fully convolutional residual network (FCRN) as the encoder and four up-projection modules as the decoder to up-sample the depth map resolution. Fu et al. \cite{fu2018deep} successfully formulated monocular depth estimation as an ordinal regression problem. Qi et al. \cite{qi2018geonet} proposed a network called GeoNet that investigate the duality between depth map and surface normal. The DNNs from Ren et al. \cite{ren2019deep} classified input images as indoor or outdoor before estimating the depth values. Lee et al. \cite{lee2019monocular} suggested an idea of using a DNNs to estimate the relative depth between pairs of pixels. The proposed method from Jiao et al. \cite{jiao2018look} incorporated object segmentation into the training to increase depth estimation accuracy. Hu et al. \cite{Hu2018RevisitingSI} introduced an architecture that included an encoder, a decoder, a multi-scale feature fusion (MFF), and a new loss term for preserving edge structures. Inspired by \cite{Hu2018RevisitingSI}, Chen et al. \cite{chen2019structure} used adaptive dense feature fusion (ADFF), and residual pyramid decoder in their network. The study by Facil et al. \cite{facil2019cam} proposed a DNNs that aims to learn calibration-aware patterns to improve the generalization capabilities of monocular depth prediction. Recently, Ramamonjisoa et al. \cite{ramamonjisoa2019sharpnet} presented SharpNet that exploits occluding contours as an additional driving factor to optimize the depth estimation model besides the depth and the surface normal. 
\\[5pt]
\noindent\textbf{Plane-based approaches:} Liu et al. \cite{liu2018planenet} was the first study to consider using the planar constraint to predict depth maps from single images. Later the same authors published an incremental study to refine the quality of plane segmentation \cite{liu2019planercnn}. Yin et al. \cite{Yin2019enforcing} formed a geometric constraint called virtual normal to predict the depth map as well as a point cloud and surface normals. Note that methods by Liu et al. focused explicitly on estimating a set of plane parameters and planar segmentation masks, while Yin et al. calculated a large virtual plane to train a DNNs that is robust to noise in the ground truth depth.
\\[5pt]
\noindent\textbf{Attention mechanism:} Attention was initially used in machine translation and it was brought to computer vision by Xu et al. \cite{xu2015show}. Since then, attention mechanism has evolved and branched into channel-wise attention \cite{hu2018squeeze,tan2019mnasnet}, spatial-wise attention \cite{bello2019attention,wang2018non} and mix attention \cite{wang2017residual} in order to tackle object detection and image classification problems. Some recent monocular depth estimation studies also followed this line of work. Xu et al. \cite{xu2018structured} proposed multi-scale spatial-wise attention to guide a Conditional Random Fields (CRFs) model. Li et al. \cite{li2018deep} proposed a discriminative depth estimation model using channel-wise attention. Kong et al. \cite{kong2019pixel} embedded a discrete binary mask, namely the pixel-wise attentional gating unit, into a residual block to modulate learned features.

In this paper, we propose using depth-attention volume (DAV) to encode non-local geometric dependencies. It can be seen as an attention mechanism that guides depth estimation to favor depth values originating from planar surfaces that are ubiquitous in man-made scenes. In contrast to previous plane-based approaches, we do not train the network to segment the planes explicitly, but instead, we let the network to learn the coplanarity constraint implicitly.

\section{Proposed Method}
This section describes the proposed depth estimation method. The first subsection defines the depth-attention volume and the following two subsections outline the network architecture and the loss functions. Further details are provided in the supplementary material.

\begin{figure}[t!]
    \includegraphics[width=0.75\textwidth]{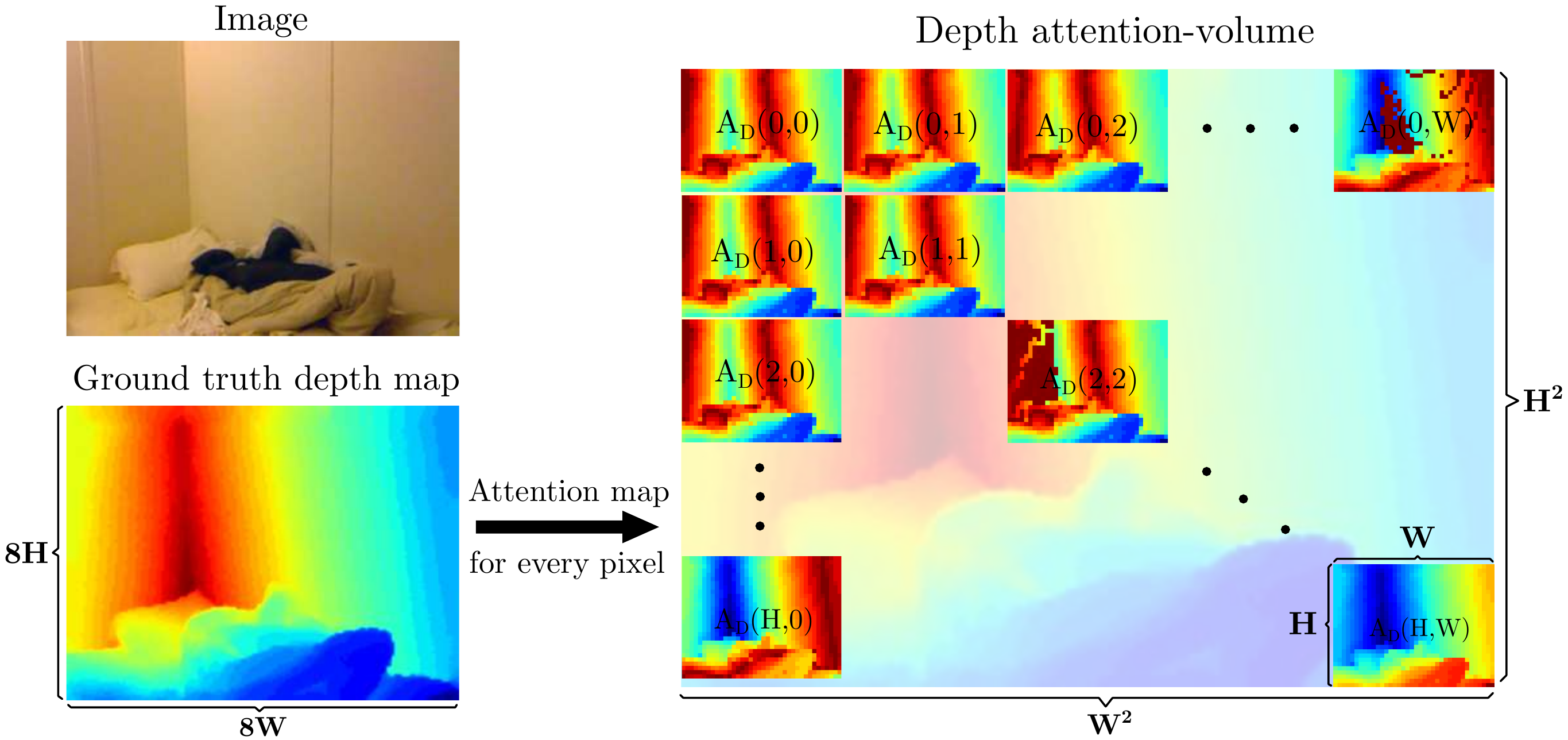}
  \centering
  \caption{Depth-attention volume (DAV) is a collection of depth-attention maps (Eq.~\ref{eq:davd}, Figure~\ref{fig:dav_illustration}) obtained using each image location as a query point at a time. Therefore, the DAV for an image of size $8H \times 8W$ is a 4D tensor of size $H \times W \times H \times W$.}
  \label{fig:dav_generation}
\end{figure}

\subsection{Depth-attention volume} \label{DAVdef}
Given two image points $P_0 = (x_0,y_0)$ and $P_1 = (x_1,y_1)$ with corresponding depth values $d_0$ and $d_1$, we define that the depth-attention $A(P_0,P_1)$ is the ability of $P_1$ to predict the depth of $P_0$. This ability is quantified as a confidence in the range $[0,1]$ so that $0$ means no ability and $1$ represents maximum certainty of being a good {\em predictor}.

To estimate $A$ we make the assumption that the scene contains multiple non-parallel planes, which is common particularly in indoor environments. The depth values of all points belonging to the same plane are linearly dependent. Hence, they are good depth predictors of each other. To exploit this property, we detect $N$ prominent planes from the training images and parameterize each plane with $S=(n_x,n_y,n_d,c)$, where $(n_x,n_y,n_d)$ is the plane normal and $c$ is the orthogonal distance from the origin. We construct the first-order depth-attention volumes for all $N$ planes:
\begin{equation}
\label{eq:dav1}    
A_i(P_0,P_1) = 1 - \sigma(|S_i \cdot X_0|+|S_i \cdot X_1|), \quad i=1,\ldots,N 
\end{equation}
where $\sigma$ is the sigmoid function, $X_0=(x_0,y_0,d_0,1)$ and $X_1=(x_1,y_1,d_1,1)$. These volumes are represented as 4-D tensors of size $H \times W \times H \times W$, where $H$ and $W$ are the vertical and horizontal sizes, respectively. In practice, one needs to subsample the volumes to keep the memory requirements reasonable. In all our experiments, we used a subsampling factor of 8.

In addition, we assume that all points located on the same fronto-parallel plane are good depth predictors of each other, because they share the same depth value. We use the ground-truth depths, and create a zero-order depth-attention volume (DAV) for every training image
\begin{equation}
\label{eq:dav0}    
A_0(P_0,P_1) = 1 - \sigma(|d_0 - d_1|). 
\end{equation}

Finally, we combine these volumes by taking the maximum attention value of all volumes:
\begin{equation}
\label{eq:davd}    
A_D(P_0,P_1) = \max(A_i(P_0,P_1)), \quad i=0,\ldots,N 
\end{equation}
It is easy to observe that DAV is a symmetric function, i.e. $A_D(P_0,P_1)=A_D(P_1,P_0)$.

If we consider $P_0$ to be a query point in the image as illustrated in Figure~\ref{fig:dav_illustration} (left), we can visualize the DAV as a two-dimensional {\em attention map} shown in Figure~\ref{fig:dav_illustration} (right). Figure~\ref{fig:dav_generation} provides an example of a depth-attention volume generated from the ground truth depth map.

\begin{figure}[b]
    \includegraphics[width=0.9\textwidth]{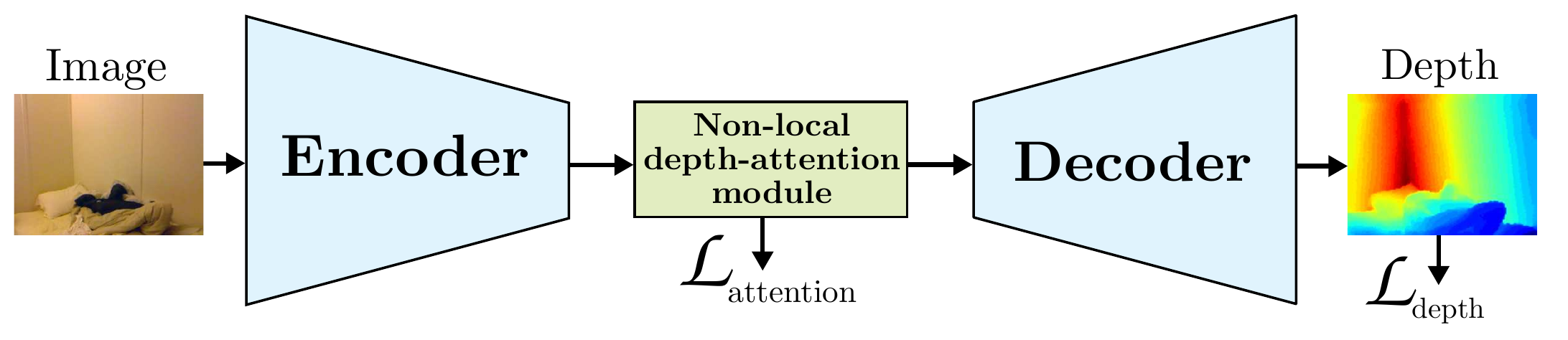}
  \centering
  \caption{The pipeline of our proposed network. An image is passed through the encoder, then the non-local depth-attention module, and finally the decoder to produce the estimated depth map. The model is trained using $\mathcal{L}_{attention}$ and $\mathcal{L}_\textit{{\mbox{depth}}}$ losses, which are described in Subsection~\ref{LossDef}.}
  \label{fig:architecture_overview}
\end{figure}

\subsection{Network Architecture} \label{NetworkDef} 

Figure~\ref{fig:architecture_overview} gives an overview of our model that includes three main modules: an encoder, a non-local depth-attention module, and a decoder. 

We opt to use a simplified dilated residual networks (DRN) with 22 layers (DRN-D-22) \cite{Yu2016,yu2017dilated} as our encoder, which extracts high-resolution features and downsamples the input image only 8 times. The DRN-D-22 is a variation of DRN that completely removes max-pooling layers as well as smoothly distributes the dilation to minimize the gridding artifacts. This is crucial to our network, because to make training feasible, the non-local depth-attention module needs to operate on a sub-sampled feature space. However, to capture meaningful spatial relationships this feature space also needs to be large enough.

The decoder part of our network contains a straightforward up-scaling scheme that increases the spatial dimension from $29 \times 38$ to $57 \times 76$ and then to $114 \times 152$. Upsampling consists of two bilinear interpolation layers followed by convolutional layers with a kernel size of $3 \times 3$. Two convolutional layers with a kernel size of $5 \times 5$ are then used to estimate the final depth map.

\begin{figure}[t!]
    \includegraphics[width=0.89\textwidth]{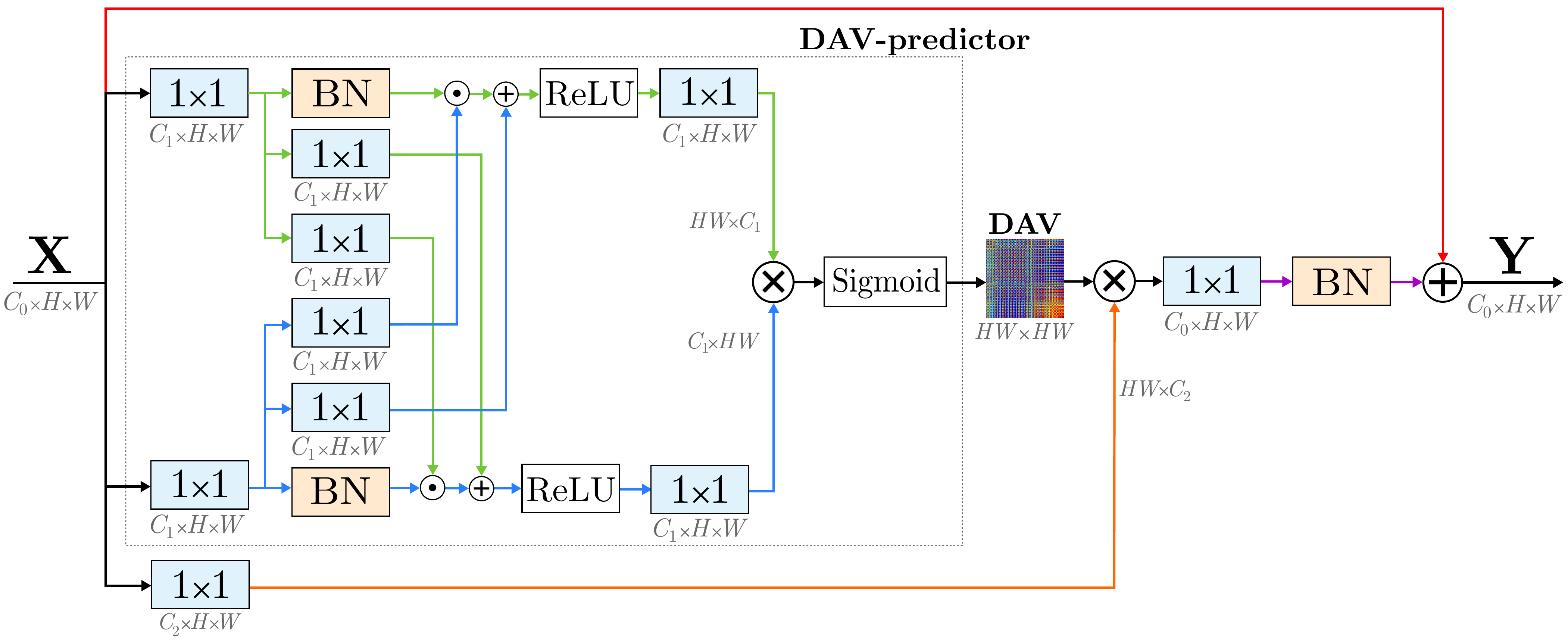}
  \centering
  \hspace{0.01mm}
  \caption{Structure of the non-local depth-attention module. ``$\bigodot$" presents element-wise multiplication, ``$\bigoplus$" presents element-wise sum, and ``$\bigotimes$" is the outer product.}
  \label{fig:dav_module}
\end{figure}

The non-local depth-attention module is located between the encoder and the decoder. It maps the input features \textbf{X} to the output features \textbf{Y} of the same size. The primary purpose of the module is to add the non-local information embedded in the depth-attention volume (DAV) to \textbf{Y}, but it is also used to predict and learn the DAV based on the ground-truth data. The structure of the module is presented in Figure~\ref{fig:dav_module}.

We implement the DAV-predictor by first transforming \textbf{X} into green and blue embeddings using $1 \times 1$ convolution. We exploit the symmetry of DAV, and maximize the correlation between these two spaces by applying cross-denormalization on both green and blue embeddings. Cross-denormalization is a conditional normalization technique \cite{dumoulin2016learned} that is used to learn an affine transformation from the data. Specifically, the green embedding is first normalized to zero mean and unit standard deviation using batch-normalization (BN). Then, the blue embedding is convolved to create two tensors that are multiplied and added the normalized features from the green branch, and vise versa. The denormalized representations are then activated with ReLUs and transformed by another $1 \times 1$ convolution before multiplying with each others. Finally, the DAV is activated using the sigmoid function to ensure that the output values are in range [0, 1]. We empirically verified that applying cross-modulation in two embedding spaces is superior than using a single embedding with double the number of features.

Furthermore, \textbf{X} is fed into the orange branch and multiplied with the estimated DAV to amplify the effect of the input features. Finally, we add a residual connection (red) to prevent the vanishing gradient problem when training our network.

\subsection{Loss Function} \label{LossDef}
As illustrated in Figure~\ref{fig:architecture_overview} our loss function consists of two main components: attention loss and depth loss.

\noindent \textbf{Attention loss:} The primary goal of this term is to minimize the error between the estimated (output of the DAV-predictor in Figure~\ref{fig:dav_module}) and the ground-truth DAV. The $\mathcal{L}_{mae}$ is defined as the mean absolute error between the predicted and the ground truth depth-attention values: 
\begin{equation} 
\label{eq:loss_atten}
\mathcal{L}_{mae} = \frac{1}{(H W)^2} \sum_{i}\sum_{j}{|\hat{A}_{i,j} - A_{i,j}|}
\end{equation}
where $\hat{A}_{i,j}\equiv \hat{A}_D(P_i, P_j)$ and $A_{i,j}\equiv A_D(P_i, P_j)$ are the predicted and ground truth depth-attention volumes.

In addition, we minimize the angle between the predicted and the ground truth depth-attention maps for all query positions $i$ and $j$:
\begin{equation} \label{eq:loss_cosine_similarity}
\mathcal{L}_{ang} = \frac{1}{H W}\left(\sum_{i}{ \left| 1 - \frac{\sum_{j}\hat{A}_{i,j}A_{i,j}}{\sqrt{\sum_{j}\hat{A}_{i,j}^2\sum_{j}A_{i,j}^2}} \right|} + \sum_{j}{ \left| 1 - \frac{\sum_{i}\hat{A}_{i,j}A_{i,j}}{\sqrt{\sum_{i}\hat{A}_{i,j}^2\sum_{i}A_{i,j}^2}} \right| }\right )
\end{equation}

The full attention loss is defined by

\begin{equation} 
\label{eq:loss_guided_depth_attention}
\mathcal{L}_{attention} = \mathcal{L}_{mae} + \lambda \mathcal{L}_{ang}
\end{equation}
where $\lambda \in \mathbb{R}^+$ is the weight loss coefficient.\\

\noindent \textbf{Depth loss:} Moreover, we define depth loss as a combination of three terms $\mathcal{L}_{log}$, $\mathcal{L}_{grad}$ and $\mathcal{L}_{norm}$ that were originally introduced in \cite{Hu2018RevisitingSI}. The $\mathcal{L}_{log}$ loss is a variation of the $L_1$ norm that is calculated in the logarithm space and defined as
\begin{equation} 
\label{eq:loss_depth_l1}
\mathcal{L}_{log} = \frac{1}{M} \sum_{i=1}^{M}{F( |\hat{d}_i - d_i| )}
\end{equation}
where $M$ is the number of valid depth values, $d_i$ is the ground truth depth, $\hat{d}_i$ is the predicted depth, and $F(x)=\log(x + \alpha)$ with $\alpha$ set to $0.5$ in our experiments. 

Another loss term is $\mathcal{L}_{grad}$, which is used to penalize sudden changes of edge structures in both $x$ and $y$ directions. It is defined by

\begin{equation} 
\label{eq:loss_grad}
\mathcal{L}_{grad} = \frac{1}{M} \sum_{i=1}^{M}{F( \Delta_x(|\hat{d}_i - d_i|) ) + F( \Delta_y(|\hat{d}_i - d_i|) )}
\end{equation}
where $\Delta_x$ and $\Delta_y$ is the gradient of the error with respect to $x$ and $y$. Finally, we use $\mathcal{L}_{norm}$ to emphasize small details by minimizing the angle between the ground truth ($n_i$) and the predicted ($\hat{n}_i$) surface normals:
\begin{equation} 
\label{eq:loss_normal}
\mathcal{L}_{norm} = \frac{1}{M} \sum_{i=1}^{M}{ |1 - \hat{n}_i \cdot n_i| }.
\end{equation}
where surface normals are estimated as $n \equiv
(-\nabla_{x}(d)$, $-\nabla_{y}(d), 1)$ using Sobel filter, like \cite{Hu2018RevisitingSI}. The depth loss is then defined by
\begin{equation} 
\label{eq:loss_refine_depth}
\mathcal{L}_{depth} = \mathcal{L}_{log} + \mu \mathcal{L}_{grad} + \theta \mathcal{L}_{norm}
\end{equation}
where $\mu, \theta \in \mathbb{R}^+$ are weight loss coefficients. Our full loss is
\begin{equation} 
\label{eq:loss_total}
\mathcal{L} = \mathcal{L}_{attention} + \gamma \mathcal{L}_{depth}
\end{equation}
where $\gamma \in \mathbb{R}^+$ is a weight loss coefficient. Subsection~\ref{implementation_detail} describes in detail how the network is trained using these loss functions.

\section{Experiments}
In this section, we evaluate the performance of the proposed method by comparing it against several baselines. We start by introducing datasets, evaluation metrics, and implementation details. The last three subsections contain the comparison to the state-of-the-art, ablation studies, and a cross-dataset evaluation. Further results are available in the supplementary material.

\subsection{Datasets and evaluation metrics}

\paragraph{\bf Datasets:} We assess the proposed method using NYU-Depth-v2 \cite{silberman2012indoor} and ScanNet \cite{dai2017scannet} datasets. NYU-Depth-v2 contains $\sim120K$ RGB-D images obtained from 464 indoor scenes. From the entire dataset, we use 50K images for training and the official test set of 654 images for evaluation. ScanNet dataset comprises of 2.5 million RGB-D images acquired from 1517 scenes. For this dataset, we use the training subset of $\sim20K$ images provided by the Robust Vision Challenge 2018 \cite{robustvision2018} (ROB). Unfortunately, the ROB test set is not available, so we report the results on the Scannet official test set of $5310$ images instead. SUN-RGBD is yet another indoor dataset consisting of $\sim10K$ images collected with four different sensors. We do not use it for training, but only for cross-evaluating the pre-trained models on the test set of 5050 images.

\paragraph{\bf Evaluation metrics:} The performance is assessed using the standard metrics provided for each dataset. That is, for NYU-Depth-v2 \cite{silberman2012indoor} we calculate the mean absolute relative error (REL), root mean square error (RMS), and thresholded accuracy ($\delta_i$). For the ScanNet and SUN-RGBD dataset, we provide the mean absolute relative error (REL), mean square relative error (sqREL), scale-invariant mean squared error (SI), mean absolute error (iMAE), and root mean square error (iRMSE) of the inverse depth values. For iBims-1 benchmark~\cite{Koch18:ECS}, we compute 5 similar metrics as for NYU-Depth-v2 plus the root mean square error in log-space (log10), planarity errors ($\epsilon^{plan}, \epsilon^{orie}$), depth boundary errors ($\epsilon^{acc}, \epsilon^{comp}$), and directed depth error ($\epsilon^{0}, \epsilon^{-}, \epsilon^{+}$). Detailed definitions of the metrics are provided in the supplementary material.

\subsection{Implementation Details} \label{implementation_detail}

The proposed model is implemented with the Pytorch \cite{NEURIPS2019_9015} framework, and trained using a single Tesla-V100, batch size of 32 images, and Adam optimizer \cite{kingma2014adam} with $(\beta_1, \beta_2, \epsilon) = (0.9, 0.999, 10^{-8})$. The training process is split into three parts. During the first phase, we replace the DAV-predictor (Figure~\ref{fig:dav_module}) with the DAVs computed from the ground truth depth maps. We train the model for 200 epochs using only the depth loss (Eq.~\ref{eq:loss_refine_depth}) and the learning rate of $10^{-4}$. In the second phase, we add the DAV-predictor to the model, freeze the weights of other parts of the model, and train for 200 epochs with the learning rate of $7.0 \times 10^{-5}$. In the last phase, we train the entire model for 300 epochs using the learning rate of $7.0 \times 10^{-5}$ for the first 100 epochs and then reduce it at the rate of $5\%$ per $25$ epochs. The last two stages employ the full loss function defined in Equation (\ref{eq:loss_total}). We set all the weight loss coefficients $\lambda, \mu, \theta$, and $\gamma$ as $1$.

We augment the training data using random scaling ([0.875, 1.25]), rotation ([-5.0, +5.0] degrees), horizontal flip, rectangular window droppings, and colorization. Planes, required for training, are obtained by fitting a parametric model to the back-projected 3D point cloud using RANSAC with the inlier threshold of $1$ cm. We select at most the best N-planes in terms of the inlier count with a maximum of 100 iterations. Furthermore, we keep only planes that cover more than $7\%$ of the image area.

\begin{table}[t!]
\caption{\label{tab:eval_nyuv2}Evaluation results on the NYU-Depth-v2 dataset. Metrics with $\downarrow$ mean lower is better and $\uparrow$ mean higher is better. Timing is the average over 1000 images using a NVIDIA GTX-1080 GPU, in frames-per-second (FPS).}
\centering
\small
\scalebox{0.89}{%
\begin{tabular}{|l|r|r|r|c|c|c|c|c|}
\hline
\textbf{Methods} & \textbf{\#params} & \textbf{Memory} & \textbf{FPS} & \textbf{REL$\downarrow$} & \textbf{RMS$\downarrow$} & \(\boldsymbol{\delta_1}\)$\uparrow$ & \(\boldsymbol{\delta_2}\)$\uparrow$ & \(\boldsymbol{\delta_3}\)$\uparrow$ \\ \hline
Eigen'15 \cite{eigen2015predicting}$^{\star\star}$ & 141.1M & - & - & 0.215 & 0.907 & 0.611 & 0.887 & 0.971 \\ \hline
Laina'16 \cite{laina2016deeper}$^{\star\star}$ & 63.4M & - & - & 0.127 & 0.573 & 0.811 & 0.953 & 0.988 \\ \hline
Liu'18 \cite{liu2018planenet}$^{\ddagger}$ & 47.5M & 124.6MB & 93 & 0.142 & 0.514 & 0.812 & 0.957 & 0.989 \\ \hline
Fu'18 \cite{fu2018deep} $^{\star\star}$ & 110.0M & 489.1MB & 42 & 0.115 & 0.509 & 0.828 & 0.965 & 0.992 \\ \hline
Qi'18 \cite{qi2018geonet} & 67.2M & - & - & 0.128 & 0.569 & 0.834 & 0.960 & 0.990 \\ \hline
Hao'18 \cite{hao2018detail} & 60.0M & - & - & 0.127 & 0.555 & 0.841 & 0.966 & 0.991 \\ \hline
Lee'19 \cite{lee2019monocular} & 118.6M & - & - & 0.131 & 0.538 & 0.837 & 0.971 & 0.994 \\ \hline
Ren'19 \cite{ren2019deep} $^{\star\star}$ & 49.8M & - & - & 0.113 & 0.501 & 0.833 & 0.968 & 0.993 \\ \hline
Zhang'19 \cite{zhang2019pattern} & 95.4M & - & - & 0.121 & 0.497 & 0.846 & 0.968 & 0.994 \\ \hline
Ramam.'19 \cite{ramamonjisoa2019sharpnet}$^{\ddagger}$ & 80.4M & 336.6MB & 47 & 0.139 & 0.502 & 0.836 & 0.966 & 0.993 \\ \hline
Hu'19 \cite{Hu2018RevisitingSI} & 157.0M & 679.7MB & 15 & 0.115 & 0.530 & 0.866 & 0.975 & 0.993 \\ \hline
Chen'19 \cite{chen2019structure} & 210.3M & 1250.9MB & 12 & 0.111 & 0.514 & 0.878 & 0.977 & 0.994 \\ \hline
Yin'19 \cite{Yin2019enforcing} & 114.2M & 437.6MB & 37 & \textbf{0.108} & 0.416 & 0.875 & 0.976 & 0.994 \\ \hline
\textbf{Ours} & \textbf{25.1M} & \textbf{96.1MB} & \textbf{218} & \textbf{0.108} & \textbf{0.412} & \textbf{0.882} & \textbf{0.980} & \textbf{0.996} \\ \hline
\end{tabular} }
\end{table}

\begin{figure}[b!]
    \includegraphics[width=0.83\textwidth]{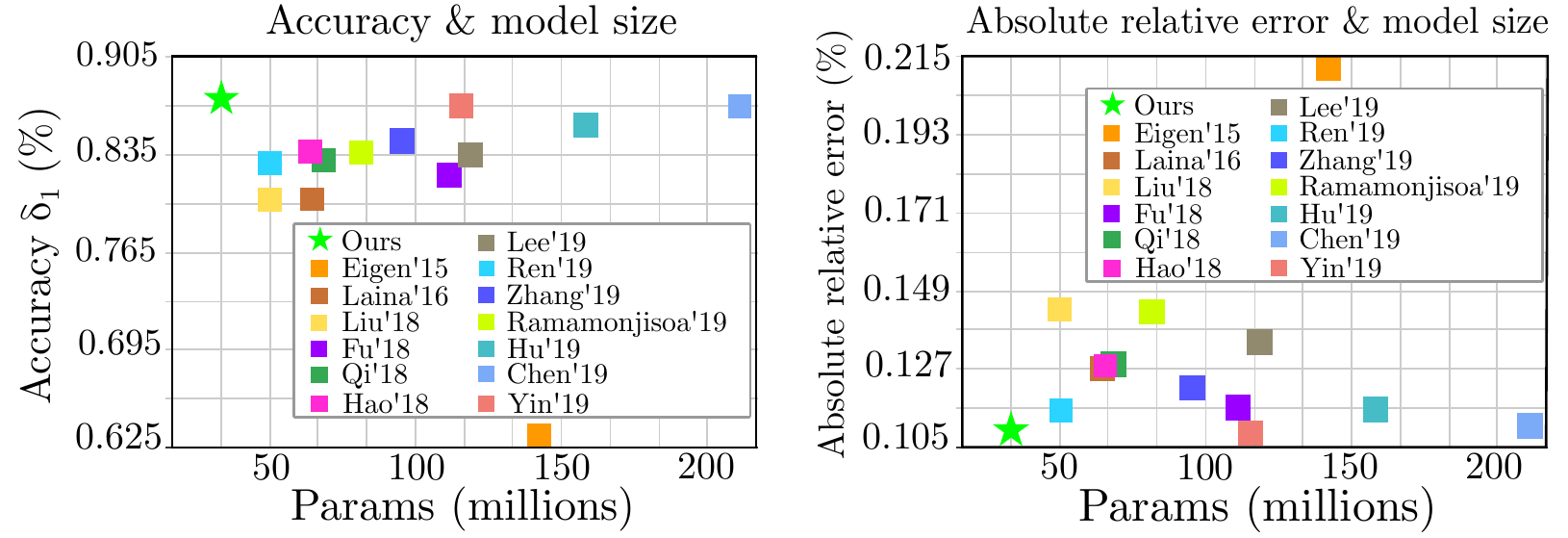}
  \centering
  \hspace{0.01mm}
  \caption{Analyzing the accuracy $\delta_1 (\%)$ and mean absolute relative error $(\%)$ with respect to the number of parameters $(millions)$ for recent monocular depth estimation methods on NYU-Depth-v2. The left picture presents the thresholded accuracy where higher values are better, while the right picture shows the absolute relative error where lower values are better.}
  \label{fig:model_params_chart}
\end{figure}

\subsection{Comparison with the state-of-the-art}
In this section, we compare the proposed approach with the current state-of-the-art monocular depth estimation methods. 

\paragraph{\bf NYU-Depth-v2:} Table~\ref{tab:eval_nyuv2} contains the performance metrics on the official NYU-Depth-v2 test set for our method and for \cite{chen2019structure,eigen2015predicting,fu2018deep,hao2018detail,Hu2018RevisitingSI,laina2016deeper,lee2019monocular,liu2018planenet,qi2018geonet,ramamonjisoa2019sharpnet,ren2019deep,Yin2019enforcing,zhang2019pattern}. In addition, the table shows the number of model parameters for each method. The performance figures for the baselines are obtained using the pre-trained models provided by the authors \cite{chen2019structure,fu2018deep,Hu2018RevisitingSI,liu2018planenet,ramamonjisoa2019sharpnet,Yin2019enforcing} or from the original papers if the model was not available \cite{eigen2015predicting,hao2018detail,laina2016deeper,lee2019monocular,qi2018geonet,ren2019deep,zhang2019pattern}. Methods indicated with $^{\star\star}$ and $^{\ddagger}$ are trained using the entire training set of 120K images or with external data, respectively. For instance, Ramamonjisoa et al. \cite{ramamonjisoa2019sharpnet} trained the method using  synthetic dataset PBRS \cite{zhang2016physically} before fine-tuning on NYU-Depth-v2. The best performance is achieved by the proposed model that also contains the least amount of parameters. The best performing baselines, Yin et al. \cite{Yin2019enforcing}, Hu et al. \cite{Hu2018RevisitingSI}, and Chen et al. \cite{chen2019structure}, have $4.5$, $6.2$, and $8.3$ times more parameters compared to ours, respectively. Figure~\ref{fig:model_params_chart} provides an additional illustration of the model parameters with respect to the performance. 

\begin{figure*}[t!]
  \centering
\hspace{-1cm}\begin{tabular}{@{} r @{\hspace{0.9\tabcolsep}} r @{\hspace{0.9\tabcolsep}} r @{\hspace{0.9\tabcolsep}} r @{\hspace{0.9\tabcolsep}} r @{\hspace{0.9\tabcolsep}} r @{\hspace{0.9\tabcolsep}} m{1em} @{}}

\raisebox{-.4\height}{\includegraphics[width=0.129\textwidth]{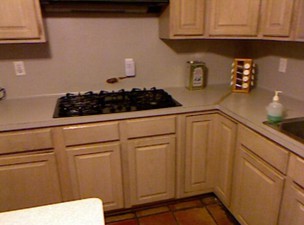}}&
\raisebox{-.4\height}{\includegraphics[width=0.129\textwidth]{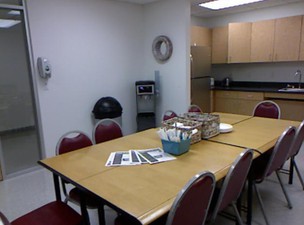}}&
\raisebox{-.4\height}{\includegraphics[width=0.129\textwidth]{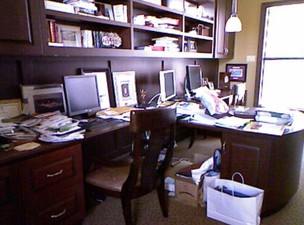}}&
\raisebox{-.4\height}{\includegraphics[width=0.129\textwidth]{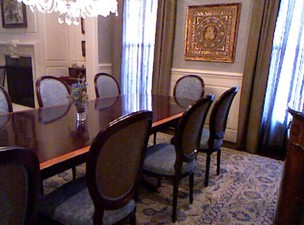}}&
\raisebox{-.4\height}{\includegraphics[width=0.129\textwidth]{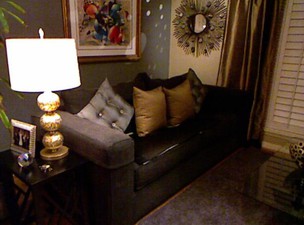}}&
\raisebox{-.4\height}{\includegraphics[width=0.129\textwidth]{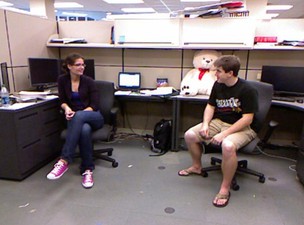}}&
Image\\
\raisebox{-.45\height}{\includegraphics[width=0.129\textwidth]{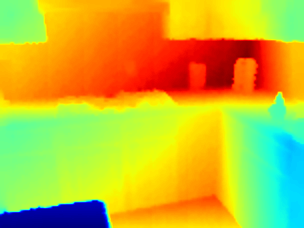}}&
\raisebox{-.45\height}{\includegraphics[width=0.129\textwidth]{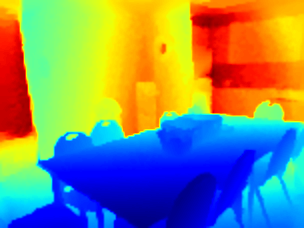}}&
\raisebox{-.45\height}{\includegraphics[width=0.129\textwidth]{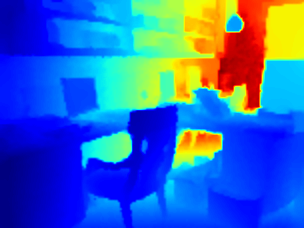}}&
\raisebox{-.45\height}{\includegraphics[width=0.129\textwidth]{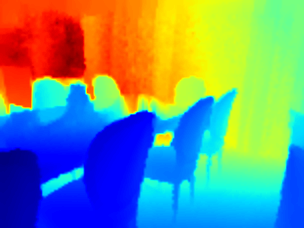}}&
\raisebox{-.45\height}{\includegraphics[width=0.129\textwidth]{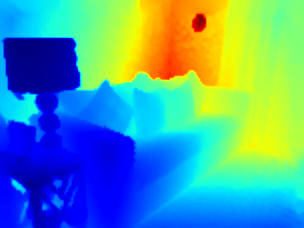}}&
\raisebox{-.45\height}{\includegraphics[width=0.129\textwidth]{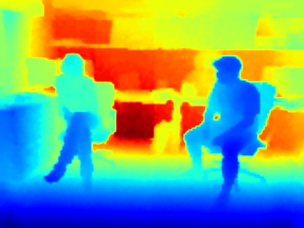}}&
Ground truth\\
\raisebox{-.45\height}{\includegraphics[width=0.129\textwidth]{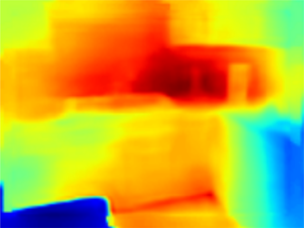}}&
\raisebox{-.45\height}{\includegraphics[width=0.129\textwidth]{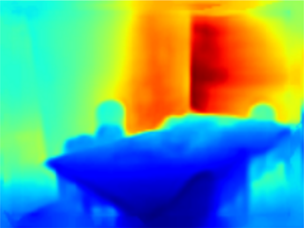}}&
\raisebox{-.45\height}{\includegraphics[width=0.129\textwidth]{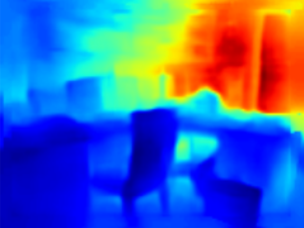}}&
\raisebox{-.45\height}{\includegraphics[width=0.129\textwidth]{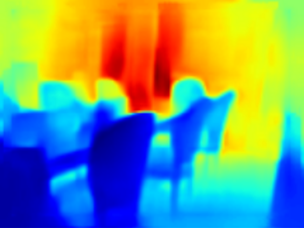}}&
\raisebox{-.45\height}{\includegraphics[width=0.129\textwidth]{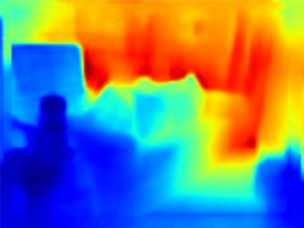}}&
\raisebox{-.45\height}{\includegraphics[width=0.129\textwidth]{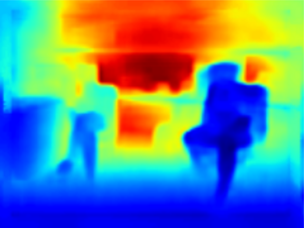}}&
Eigen'15 \cite{eigen2015predicting}\\
\raisebox{-.45\height}{\includegraphics[width=0.129\textwidth]{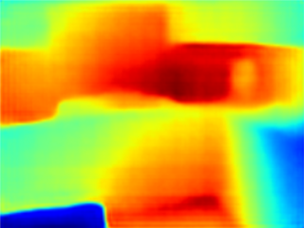}}&
\raisebox{-.45\height}{\includegraphics[width=0.129\textwidth]{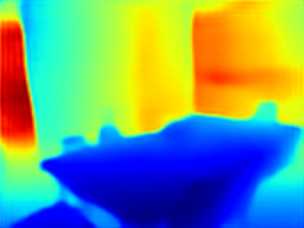}}&
\raisebox{-.45\height}{\includegraphics[width=0.129\textwidth]{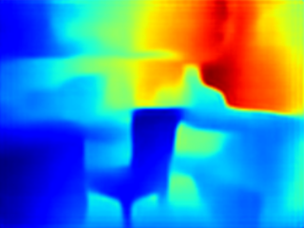}}&
\raisebox{-.45\height}{\includegraphics[width=0.129\textwidth]{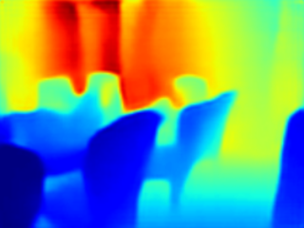}}&
\raisebox{-.45\height}{\includegraphics[width=0.129\textwidth]{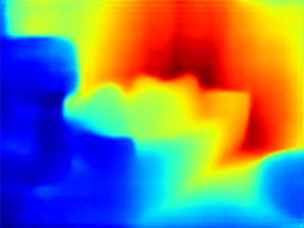}}&
\raisebox{-.45\height}{\includegraphics[width=0.129\textwidth]{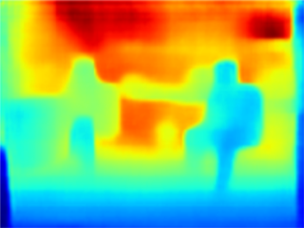}}&
Laina'16 \cite{laina2016deeper}\\
\raisebox{-.45\height}{\includegraphics[width=0.129\textwidth]{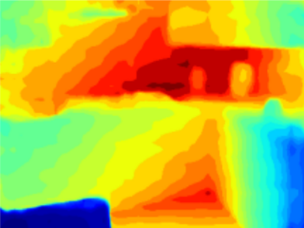}}&
\raisebox{-.45\height}{\includegraphics[width=0.129\textwidth]{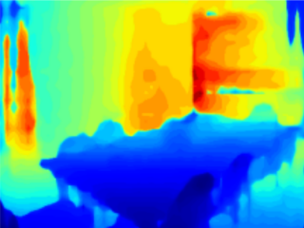}}&
\raisebox{-.45\height}{\includegraphics[width=0.129\textwidth]{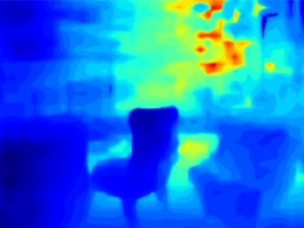}}&
\raisebox{-.45\height}{\includegraphics[width=0.129\textwidth]{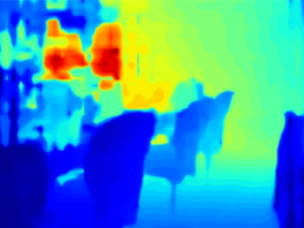}}&
\raisebox{-.45\height}{\includegraphics[width=0.129\textwidth]{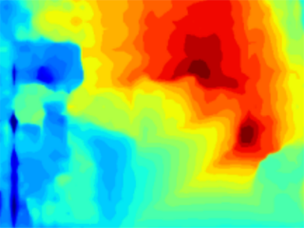}}&
\raisebox{-.45\height}{\includegraphics[width=0.129\textwidth]{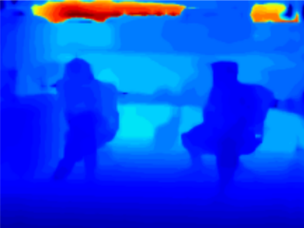}}&
Fu'18 \cite{fu2018deep}\\
\raisebox{-.45\height}{\includegraphics[width=0.129\textwidth]{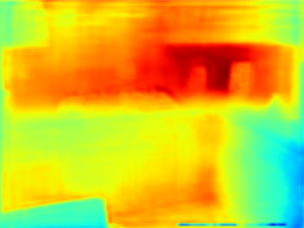}}&
\raisebox{-.45\height}{\includegraphics[width=0.129\textwidth]{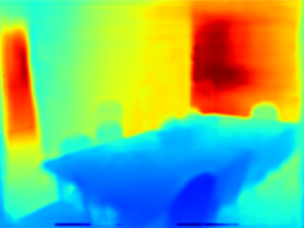}}&
\raisebox{-.45\height}{\includegraphics[width=0.129\textwidth]{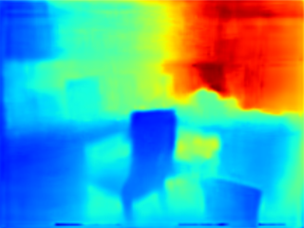}}&
\raisebox{-.45\height}{\includegraphics[width=0.129\textwidth]{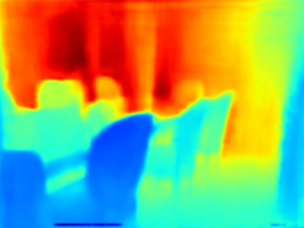}}&
\raisebox{-.45\height}{\includegraphics[width=0.129\textwidth]{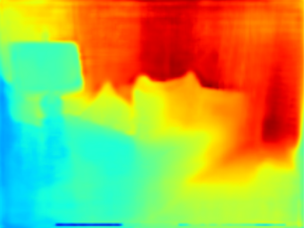}}&
\raisebox{-.45\height}{\includegraphics[width=0.129\textwidth]{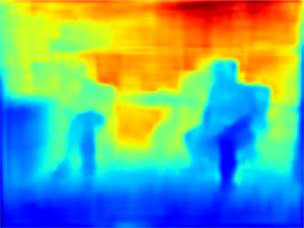}}&
Ramam.'19 \cite{ramamonjisoa2019sharpnet}\\
\raisebox{-.45\height}{\includegraphics[width=0.129\textwidth]{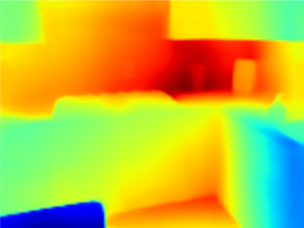}}&
\raisebox{-.45\height}{\includegraphics[width=0.129\textwidth]{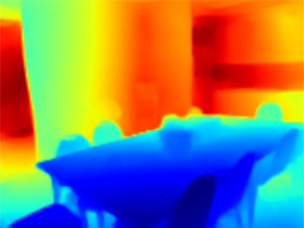}}&
\raisebox{-.45\height}{\includegraphics[width=0.129\textwidth]{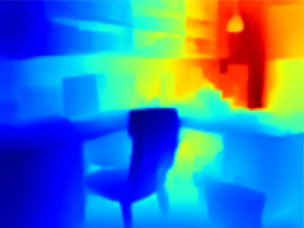}}&
\raisebox{-.45\height}{\includegraphics[width=0.129\textwidth]{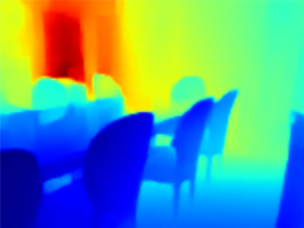}}&
\raisebox{-.45\height}{\includegraphics[width=0.129\textwidth]{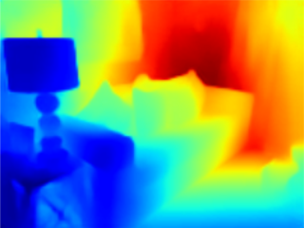}}&
\raisebox{-.45\height}{\includegraphics[width=0.129\textwidth]{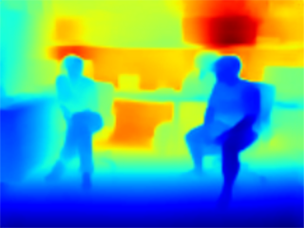}}&
Hu'19 \cite{Hu2018RevisitingSI}\\
\raisebox{-.45\height}{\includegraphics[width=0.129\textwidth]{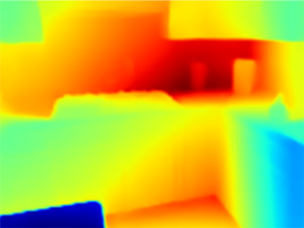}}&
\raisebox{-.45\height}{\includegraphics[width=0.129\textwidth]{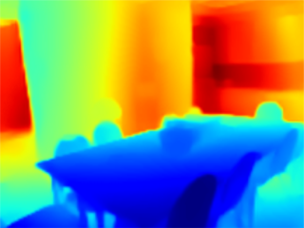}}&
\raisebox{-.45\height}{\includegraphics[width=0.129\textwidth]{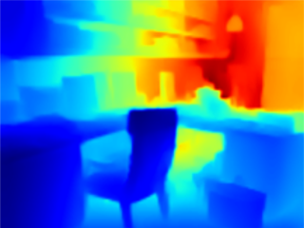}}&
\raisebox{-.45\height}{\includegraphics[width=0.129\textwidth]{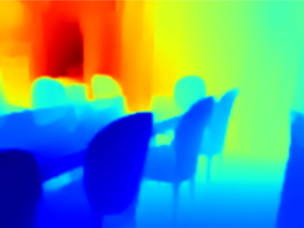}}&
\raisebox{-.45\height}{\includegraphics[width=0.129\textwidth]{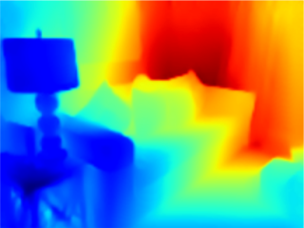}}&
\raisebox{-.45\height}{\includegraphics[width=0.129\textwidth]{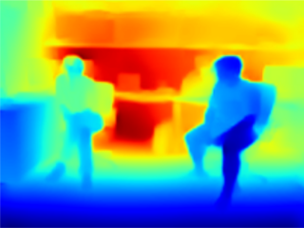}}&
Chen'19 \cite{chen2019structure}\\
\raisebox{-.45\height}{\includegraphics[width=0.129\textwidth]{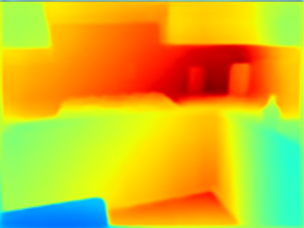}}&
\raisebox{-.45\height}{\includegraphics[width=0.129\textwidth]{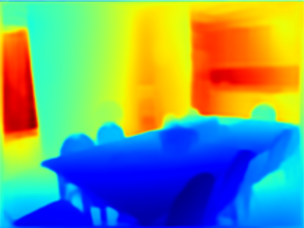}}&
\raisebox{-.45\height}{\includegraphics[width=0.129\textwidth]{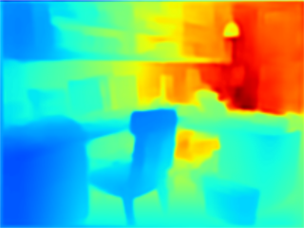}}&
\raisebox{-.45\height}{\includegraphics[width=0.129\textwidth]{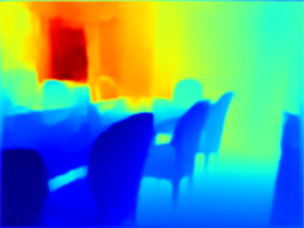}}&
\raisebox{-.45\height}{\includegraphics[width=0.129\textwidth]{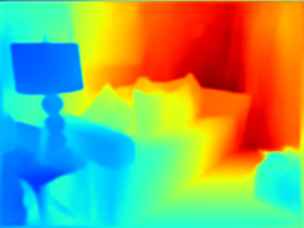}}&
\raisebox{-.45\height}{\includegraphics[width=0.129\textwidth]{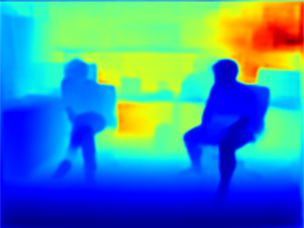}}&
Yin'19 \cite{Yin2019enforcing}\\
\raisebox{-.45\height}{\includegraphics[width=0.129\textwidth]{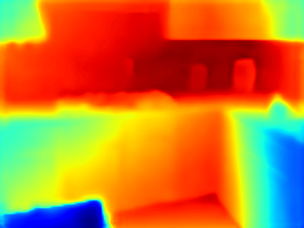}}&
\raisebox{-.45\height}{\includegraphics[width=0.129\textwidth]{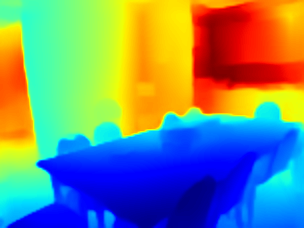}}&
\raisebox{-.45\height}{\includegraphics[width=0.129\textwidth]{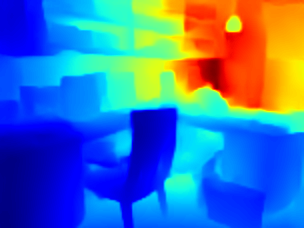}}&
\raisebox{-.45\height}{\includegraphics[width=0.129\textwidth]{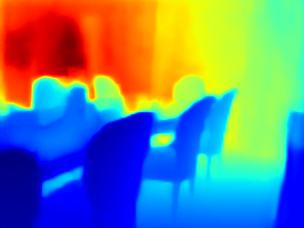}}&
\raisebox{-.45\height}{\includegraphics[width=0.129\textwidth]{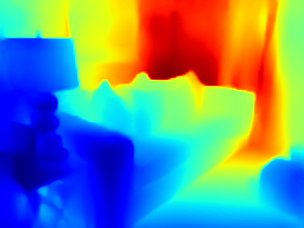}}&
\raisebox{-.45\height}{\includegraphics[width=0.129\textwidth]{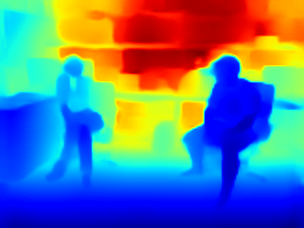}}&
Ours\\
\end{tabular}
    \caption{Qualitative results on the official NYU-Depth-v2 \cite{silberman2012indoor} test set from different methods. The color indicates the distance where red is far and blue is close. Our estimated depth maps are closer to the ground truth depth when comparing with state-of-art methods.}
    \label{fig:qualitative_nyuv2}
\end{figure*}

Figure~\ref{fig:qualitative_nyuv2} shows qualitative examples of the obtained depth maps. In this case, the maps for the baseline methods are produced using the pre-trained models provided by the authors. The method by Eigen and Fergus \cite{eigen2015predicting} performs well on uniform regions, but has difficulties in detailed structures. Laina et al. \cite{laina2016deeper} produces overly smoothed depth maps losing many small details. In contrast, Fu et al. \cite{fu2018deep} returns many details, but with the expense of discontinuities inside objects or smooth areas. The depth images by Ramamonjisoa et al. \cite{ramamonjisoa2019sharpnet} contain noise and are prone to miss fine details. Yin et al. \cite{Yin2019enforcing}, Hu et al. \cite{Hu2018RevisitingSI}, and Chen et al. \cite{chen2019structure} provide the best results among the baselines. However, they have difficulties e.g. on the third (near the desk and table) and the fourth examples from the left (wall area). We provide further qualitative examples in the supplementary material.

%Tabular description
% c: center
% l: left-justify column
% m: middle

\begin{table}[t!]
\caption{\label{tab:eval_scannet}Evaluation results on ScanNet \cite{dai2017scannet}.}
\centering
\small
\begin{tabular}{|l|r|c|c|c|c|c|c|}
\hline
\textbf{Architecture} & \textbf{\#params} & \textbf{REL} & \textbf{sqREL} & \textbf{SI} & \textbf{iMAE} & \textbf{iRMSE}  & \textbf{Test set} \\ \hline
CSWS\_E\_ROB \cite{li2018monocular} & 65.8M & 0.150 & 0.060 & 0.020 & 0.100 & 0.130 & \multirow{3}{*}{ROB} \\ \cline{1-7}
DORN\_ROB \cite{fu2018deep} & 110.0M & 0.140 & 0.060 & 0.020 & 0.100 & 0.130 & \\ \cline{1-7}
DABC\_ROB \cite{li2018deep} & 56.6M & 0.140 & 0.060 & 0.020 & 0.100 & 0.130 & \\ \hline
Hu'19 \cite{Hu2018RevisitingSI} & 157.0M & 0.139 & 0.081 & 0.016 & 0.100 & 0.105 & \multirow{4}{*}{Official} \\ \cline{1-7}
Chen'19 \cite{chen2019structure} & 210.3M & 0.134 & 0.077 & \textbf{0.015} & 0.093 & 0.100 & \\ \cline{1-7}
Ren'19 \cite{ren2019deep} & 49.8M & 0.138 & \textbf{0.057} & - & - & - & \\ \cline{1-7}
% Facil'19 \cite{facil2019cam}$^{\ddagger}$ & - & 0.120 & \textbf{0.030} & - & 0.090 & 0.120 \\ \hline
\textbf{Ours} & \textbf{25.1M} & \textbf{0.118} & \textbf{0.057} & \textbf{0.015} & \textbf{0.089} & \textbf{0.097} & \\ \hline
\end{tabular}
\end{table}

\begin{figure}[b!]
    \includegraphics[width=0.83\textwidth]{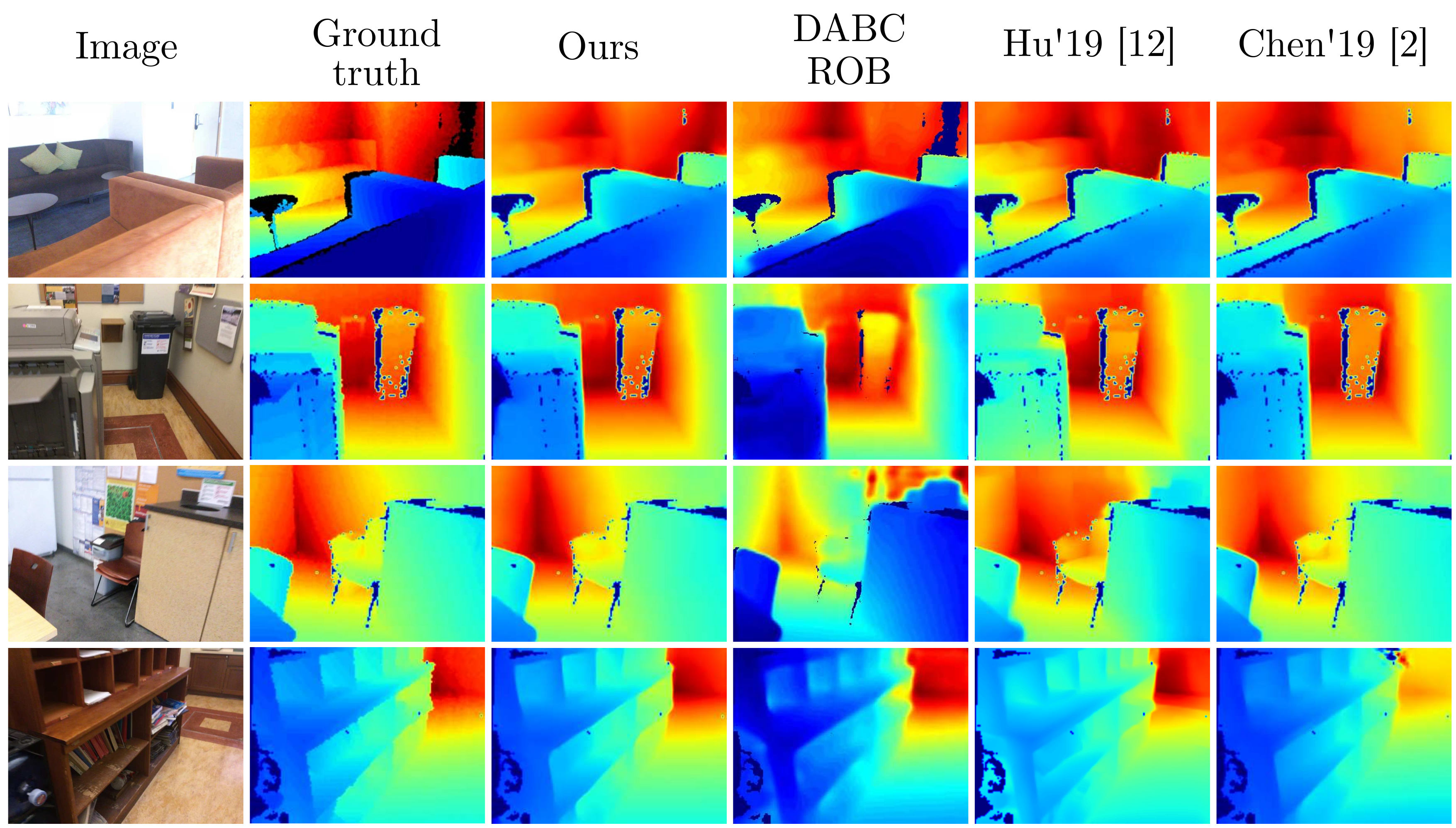}
  \centering
  \caption{Predicted depth maps from our model with baselines on the official ScanNet \cite{dai2017scannet} test set.}
  \label{fig:qualitative_scannet}
\end{figure}

\paragraph{\bf ScanNet:} 
Table~\ref{tab:eval_scannet} contains the performance figures on the official ScanNet test set for our method, Ren et al. \cite{ren2019deep} (taken from the original paper), Hu et al. \cite{Hu2018RevisitingSI} and Chen et al. \cite{chen2019structure}. We use the public code from \cite{chen2019structure,Hu2018RevisitingSI} to train their models. Unfortunately, the other baselines do not provide the results for ScanNet official test set. Moreover, the test set used in the Robust Vision Challenge (ROB) is not available at the moment and we are unable to report our performance on that. Nevertheless, we have included the best methods from the ROB challenge in Table~\ref{tab:eval_scannet} to provide indicative comparison. Note that all methods are trained with the same ROB training split. The proposed model outperforms \cite{ren2019deep} with a clear margin in terms of REL. The results are also substantially better compared to ROB challenge methods, although the comparison is not strictly fair due to different test splits. Figure~\ref{fig:qualitative_scannet} provides qualitative comparison between our method and \cite{chen2019structure,Hu2018RevisitingSI,li2018deep}, using the sample images provided in \cite{li2018deep}. The geometric structures and details are clearly better extracted by our method.

\begin{table}[b!]
    \caption{\label{tab:eval_ibims1}The iBims-1 benchmark}
    \centering
    \small
    \scalebox{0.83}{%
    \begin{tabular}{|l|c|c|c|c|c|c|c|c|c|c|c|c|c|}
    \hline
    \textbf{Method} & \textbf{REL}$\downarrow$ & \textbf{log10}$\downarrow$ & \textbf{RMS}$\downarrow$ & \(\boldsymbol{\delta_1}\)$\uparrow$ & \(\boldsymbol{\delta_2}\)$\uparrow$ & \(\boldsymbol{\delta_3}\)$\uparrow$ & \(\boldsymbol{\epsilon^{plan}}\)$\downarrow$ & \(\boldsymbol{\epsilon^{orie}}\)$\downarrow$ & \(\boldsymbol{\epsilon^{acc}}\)$\downarrow$ & \(\boldsymbol{\epsilon^{comp}}\)$\downarrow$ & \(\boldsymbol{\epsilon^{0}}\)$\uparrow$ & \(\boldsymbol{\epsilon^{-}}\)$\downarrow$ & \(\boldsymbol{\epsilon^{+}}\)$\downarrow$ \\ \hline
     Liu'18~\cite{liu2018planenet} & 0.29 & 0.17 & 1.45 & 0.41 & 0.70 & 0.86 & 7.26 & \textbf{17.24} & 4.84 & 8.86 & 71.24 & 28.36 & \textbf{0.40} \\ %
     \hline
     Ramam.'19~\cite{ramamonjisoa2019sharpnet} & 0.26 & 0.11 & 1.07 & \textbf{0.59} & \textbf{0.84} & \textbf{0.94} & 9.95 & 25.67 & 3.52 & 7.61 & 84.03 & 9.48 & 6.49 \\ %
     \hline
     \textbf{Ours} & \textbf{0.24} & \textbf{0.10} & \textbf{1.06} & \textbf{0.59} & \textbf{0.84} & \textbf{0.94} & \textbf{7.21} & 18.45 & \textbf{3.46} & \textbf{7.43} & \textbf{84.36} & \textbf{6.84} & 6.27 \\ \hline
    \end{tabular} }
\end{table}

\paragraph{\bf Planarity error analysis:}
We also evaluated our method on the iBims-1 benchmark~\cite{Koch18:ECS} and compared it with two recent works~\cite{liu2018planenet,ramamonjisoa2019sharpnet}. The results, shown in Table~\ref{tab:eval_ibims1}, indicate that we outperform the baselines in most of the metrics, including plane related ones. Extensive planarity analysis is provided in the supplementary material.

\subsection{Ablation studies}

Firstly, we assess how the number of prominent planes, used to estimate the ground truth DAVs in the training phase, affects the performance (see Sec.~\ref{DAVdef}). To this end, we train our model using the fronto-parallel planes (see Eq.~\ref{eq:dav0}) plus three, five, and seven non-fronto-parallel planes ($N$ in Eq.~\ref{eq:dav1}). The corresponding results for the NYU-Depth-v2 test set are provided in Table~\ref{tab:ablation_plane}. One can observe that the results improve by increasing the number of planes up to five and decrease after that. Possible explanation for this could be that the images used in the experiments do not typically contain more than five significant planes that can predict the depth values reliably. We also re-trained our model without the non-local depth attention (DAV) module (and any planes) and the performance degraded substantially as shown in Table~\ref{tab:ablation_plane}.

\begin{table}[b!]
\caption{\label{tab:ablation_plane}Performance of our model using different types of depth-attention volume.}
\centering
\small
\begin{tabular}{|l|c|c|c|c|c|}
\hline
\textbf{DAV-types} & \textbf{REL$\downarrow$} & \textbf{RMS$\downarrow$} & \(\boldsymbol{\delta_1}\)$\uparrow$ & \(\boldsymbol{\delta_2}\)$\uparrow$ & \(\boldsymbol{\delta_3}\)$\uparrow$ \\ \hline
w/o DAV-module & 0.140 & 0.577 & 0.827 & 0.960 & 0.989 \\ \hline
$||$-Plane-DAV & 0.116 & 0.442 & 0.867 & 0.976 & 0.995 \\ \hline
3-Plane-DAV & 0.110 & 0.421 & 0.879 & 0.978 & 0.995 \\ \hline
\textbf{5-Plane-DAV} & \textbf{0.108} & \textbf{0.412} & \textbf{0.882} & \textbf{0.980} & \textbf{0.996} \\ \hline
7-Plane-DAV & 0.111 & 0.447 & 0.851 & 0.970 & 0.993 \\ \hline
\end{tabular}
\end{table}

\begin{table}[b!]
\caption{\label{tab:ablation_w_o_l_atten}Ablation studies of models without and with the attention loss on the NYU-Depth-v2. This shows the importance of the DAV in guiding the monocular depth model.}
\centering
\small
\begin{tabular}{|l|c|c|c|c|c|}
\hline
\textbf{Training} & \textbf{REL$\downarrow$} & \textbf{RMS$\downarrow$} & \(\boldsymbol{\delta_1}\)$\uparrow$ & \(\boldsymbol{\delta_2}\)$\uparrow$ & \(\boldsymbol{\delta_3}\)$\uparrow$ \\ \hline
w/o $\mathcal{L}_{attention}$ & 0.126 & 0.540 & 0.841 & 0.967 & 0.992 \\ \hline
\textbf{w/ full loss} & \textbf{0.108} & \textbf{0.412} & \textbf{0.882} & \textbf{0.980} & \textbf{0.996} \\ \hline
continue w/o $\mathcal{L}_{attention}$ & 0.109 & 0.415 & 0.882 & 0.979 & 0.995 \\ \hline
\end{tabular}
\end{table}

Secondly, we study the impact of the attention loss term (Eq.~\ref{eq:loss_guided_depth_attention}). For this purpose, we first train our model with and without the attention loss, and then continue training by dropping the attention loss after convergence. We report the results in Table~\ref{tab:ablation_w_o_l_atten}. The model without the attention loss has clearly inferior performance indicating the importance of this loss term. Furthermore, continuing training by dropping the attention loss also degrades the performance.

\subsection{Cross-dataset evaluation} 

\begin{figure}[t]
    \includegraphics[width=0.82\textwidth]{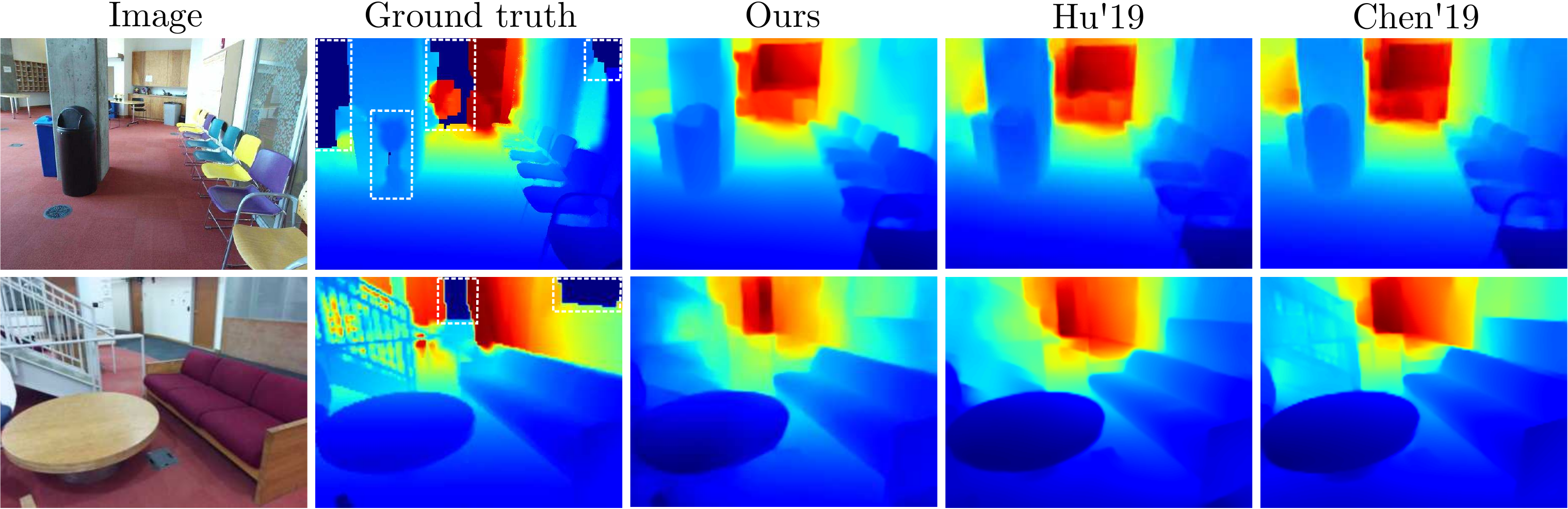}
  \centering
  \caption{Direct results on SUN RGB-D dataset \cite{song2015sun} without fine-tuning. Some regions in the white boxes show missing or incorrect depth values from the ground truth data.}
  \label{fig:generalization}
\end{figure}

To assess the generalisation properties of the model, we perform a cross-dataset evaluation, where we train the network using NYU-Depth-v2 and test with SUN-RGBD \cite{janoch2013category,song2015sun,xiao2013sun3d} without any fine-tuning. We also evaluate the baseline methods from \cite{chen2019structure,Hu2018RevisitingSI} and report the results in Table~\ref{tab:eval_generalization}. As can be seen our model performs favourably compared to the other methods. Figure~\ref{fig:generalization} contains a few examples of the results with the SUN-RGBD dataset. One can observe that our model is able to well estimate the geometric structures and details of the scene despite the differences in data distributions between the training and testing sets. Moreover, we evaluated our model without the DAV-module in the same cross-dataset setup. The results, shown in Table~\ref{tab:eval_generalization}, clearly demonstrates that the DAV-module improves the generalization.

\begin{table}[t!]
\caption{\label{tab:eval_generalization}Cross-dataset evaluation with training on NYU-Depth-v2 and testing on SUN-RGBD.}
\centering
\small
\begin{tabular}{|l|r|c|c|c|c|c|}
\hline
\textbf{Models} & \textbf{\#params} & \textbf{REL} & \textbf{sqREL} & \textbf{SI} & \textbf{iMAE} & \textbf{iRMSE} \\ \hline
 w/o DAV-module & \textbf{17.5M} & 0.254 & 0.416 & 0.035 & 0.111 & 0.091 \\ \hline
 Hu'19 \cite{Hu2018RevisitingSI} & 157.0M & 0.245 & 0.389 & 0.031 & 0.108 & 0.087 \\
 \hline
 Chen'19 \cite{chen2019structure} & 210.3M & 0.243 & 0.393 & 0.031 & \textbf{0.102} & \textbf{0.069} \\
 \hline
 \textbf{Ours} & \textbf{25.1M} & \textbf{0.238} & \textbf{0.387} & \textbf{0.030} & 0.104 & 0.075 \\ \hline
\end{tabular}
\end{table}

\section{Conclusions}
This paper proposed a novel monocular depth estimation method that incorporates a non-local coplanarity constraint with a novel attention mechanism called depth-attention volume (DAV). The proposed attention mechanism encourages depth estimation to favor planar structures, which are common especially in indoor environments. The DAV enables more efficient learning of the necessary priors, which results in considerable reduction in the number of model parameters. The performance of the proposed solution is state-of-the-art on two popular benchmark datasets while using 2-8 times less parameters than competing methods. Finally, the generalisation ability of the method was further demonstrated in cross dataset experiments. 

\clearpage

\bibliographystyle{splncs04}
\bibliography{ms}

\begin{thebibliography}{10}
\providecommand{\url}[1]{\texttt{#1}}
\providecommand{\urlprefix}{URL }
\providecommand{\doi}[1]{https://doi.org/#1}

\bibitem{chen2019structure_supp}
Chen, X., Chen, X., Zha, Z.J.: Structure-aware residual pyramid network for
  monocular depth estimation. In: Proceedings of the 28th International Joint
  Conference on Artificial Intelligence. pp. 694--700. AAAI Press (2019)

\bibitem{dai2017scannet_supp}
Dai, A., Chang, A.X., Savva, M., Halber, M., Funkhouser, T., Nie{\ss}ner, M.:
  Scannet: Richly-annotated 3d reconstructions of indoor scenes. In:
  Proceedings of the IEEE Conference on Computer Vision and Pattern
  Recognition. pp. 5828--5839 (2017)

\bibitem{deng2009imagenet_supp}
Deng, J., Dong, W., Socher, R., Li, L.J., Li, K., Fei-Fei, L.: Imagenet: A
  large-scale hierarchical image database. In: 2009 IEEE conference on computer
  vision and pattern recognition. pp. 248--255. Ieee (2009)

\bibitem{eigen2015predicting_supp}
Eigen, D., Fergus, R.: Predicting depth, surface normals and semantic labels
  with a common multi-scale convolutional architecture. In: Proceedings of the
  IEEE International Conference on Computer Vision. pp. 2650--2658 (2015)

\bibitem{eigen2014depth_supp}
Eigen, D., Puhrsch, C., Fergus, R.: Depth map prediction from a single image
  using a multi-scale deep network. In: Advances in neural information
  processing systems. pp. 2366--2374 (2014)

\bibitem{hao2018detail_supp}
Hao, Z., Li, Y., You, S., Lu, F.: Detail preserving depth estimation from a
  single image using attention guided networks. In: 2018 International
  Conference on 3D Vision (3DV). pp. 304--313. IEEE (2018)

\bibitem{Hu2018RevisitingSI_supp}
Hu, J., Ozay, M., Zhang, Y., Okatani, T.: Revisiting single image depth
  estimation: Toward higher resolution maps with accurate object boundaries.
  In: IEEE Winter Conf. on Applications of Computer Vision (WACV) (2019)

\bibitem{janoch2013category_supp}
Janoch, A., Karayev, S., Jia, Y., Barron, J.T., Fritz, M., Saenko, K., Darrell,
  T.: A category-level 3d object dataset: Putting the kinect to work. In:
  Consumer Depth Cameras for Computer Vision, pp. 141--165. Springer (2013)

\bibitem{Koch18:ECS_supp}
Koch, T., Liebel, L., Fraundorfer, F., K{\"o}rner, M.: Evaluation of cnn-based
  single-image depth estimation methods. In: Leal-Taixé, L., Roth, S. (eds.)
  European Conference on Computer Vision Workshop (ECCV-WS). pp. 331--348.
  Springer International Publishing (2018)

\bibitem{laina2016deeper_supp}
Laina, I., Rupprecht, C., Belagiannis, V., Tombari, F., Navab, N.: Deeper depth
  prediction with fully convolutional residual networks. In: 2016 Fourth
  International Conference on 3D Vision (3DV). pp. 239--248. IEEE (2016)

\bibitem{li2017two_supp}
Li, J., Klein, R., Yao, A.: A two-streamed network for estimating fine-scaled
  depth maps from single rgb images. In: Proceedings of the IEEE International
  Conference on Computer Vision. pp. 3372--3380 (2017)

\bibitem{liu2018planenet_supp}
Liu, C., Yang, J., Ceylan, D., Yumer, E., Furukawa, Y.: Planenet: Piece-wise
  planar reconstruction from a single rgb image. In: Proceedings of the IEEE
  Conference on Computer Vision and Pattern Recognition. pp. 2579--2588 (2018)

\bibitem{liu2015deep_supp}
Liu, F., Shen, C., Lin, G.: Deep convolutional neural fields for depth
  estimation from a single image. In: Proceedings of the IEEE conference on
  computer vision and pattern recognition. pp. 5162--5170 (2015)

\bibitem{ramamonjisoa2019sharpnet_supp}
Ramamonjisoa, M., Lepetit, V.: Sharpnet: Fast and accurate recovery of
  occluding contours in monocular depth estimation. The IEEE International
  Conference on Computer Vision (ICCV) Workshops  (2019)

\bibitem{ren2019deep_supp}
Ren, H., El-khamy, M., Lee, J.: Deep robust single image depth estimation
  neural network using scene understanding. In: Proceedings of the IEEE
  Conference on Computer Vision and Pattern Recognition Workshops. pp. 37--45
  (2019)

\bibitem{silberman2012indoor_supp}
Silberman, N., Hoiem, D., Kohli, P., Fergus, R.: Indoor segmentation and
  support inference from rgbd images. In: European Conference on Computer
  Vision. pp. 746--760. Springer (2012)

\bibitem{song2015sun_supp}
Song, S., Lichtenberg, S.P., Xiao, J.: Sun rgb-d: A rgb-d scene understanding
  benchmark suite. In: Proceedings of the IEEE conference on computer vision
  and pattern recognition. pp. 567--576 (2015)

\bibitem{xiao2013sun3d_supp}
Xiao, J., Owens, A., Torralba, A.: Sun3d: A database of big spaces
  reconstructed using sfm and object labels. In: Proceedings of the IEEE
  International Conference on Computer Vision. pp. 1625--1632 (2013)

\bibitem{Yin2019enforcing_supp}
Yin, W., Liu, Y., Shen, C., Yan, Y.: Enforcing geometric constraints of virtual
  normal for depth prediction. In: The IEEE International Conference on
  Computer Vision (ICCV) (2019)

\bibitem{Yu2016_supp}
Yu, F., Koltun, V.: Multi-scale context aggregation by dilated convolutions.
  In: International Conference on Learning Representations (ICLR) (2016)

\bibitem{yu2017dilated_supp}
Yu, F., Koltun, V., Funkhouser, T.: Dilated residual networks. In: Proceedings
  of the IEEE conference on computer vision and pattern recognition. pp.
  472--480 (2017)

\bibitem{zhang2019pattern_supp}
Zhang, Z., Cui, Z., Xu, C., Yan, Y., Sebe, N., Yang, J.: Pattern-affinitive
  propagation across depth, surface normal and semantic segmentation. In:
  Proceedings of the IEEE Conference on Computer Vision and Pattern
  Recognition. pp. 4106--4115 (2019)

\end{thebibliography}


\begin{thebibliography}{10}
\providecommand{\url}[1]{\texttt{#1}}
\providecommand{\urlprefix}{URL }
\providecommand{\doi}[1]{https://doi.org/#1}

\bibitem{bello2019attention}
Bello, I., Zoph, B., Vaswani, A., Shlens, J., Le, Q.V.: Attention augmented
  convolutional networks. In: Proceedings of the IEEE International Conference
  on Computer Vision. pp. 3286--3295 (2019)

\bibitem{chen2019structure}
Chen, X., Chen, X., Zha, Z.J.: Structure-aware residual pyramid network for
  monocular depth estimation. In: Proceedings of the 28th International Joint
  Conference on Artificial Intelligence. pp. 694--700. AAAI Press (2019)

\bibitem{dai2017scannet}
Dai, A., Chang, A.X., Savva, M., Halber, M., Funkhouser, T., Nie{\ss}ner, M.:
  Scannet: Richly-annotated 3d reconstructions of indoor scenes. In:
  Proceedings of the IEEE Conference on Computer Vision and Pattern
  Recognition. pp. 5828--5839 (2017)

\bibitem{dumoulin2016learned}
Dumoulin, V., Shlens, J., Kudlur, M.: A learned representation for artistic
  style. {I}n: {I}nternational {C}onference on {L}earning {R}epresentations
  {ICLR}  (2017)

\bibitem{eigen2015predicting}
Eigen, D., Fergus, R.: Predicting depth, surface normals and semantic labels
  with a common multi-scale convolutional architecture. In: Proceedings of the
  IEEE International Conference on Computer Vision. pp. 2650--2658 (2015)

\bibitem{eigen2014depth}
Eigen, D., Puhrsch, C., Fergus, R.: Depth map prediction from a single image
  using a multi-scale deep network. In: Advances in neural information
  processing systems. pp. 2366--2374 (2014)

\bibitem{facil2019cam}
Facil, J.M., Ummenhofer, B., Zhou, H., Montesano, L., Brox, T., Civera, J.:
  Cam-convs: camera-aware multi-scale convolutions for single-view depth. In:
  Proceedings of the IEEE Conference on Computer Vision and Pattern
  Recognition. pp. 11826--11835 (2019)

\bibitem{fu2018deep}
Fu, H., Gong, M., Wang, C., Batmanghelich, K., Tao, D.: Deep ordinal regression
  network for monocular depth estimation. In: Proceedings of the IEEE
  Conference on Computer Vision and Pattern Recognition. pp. 2002--2011 (2018)

\bibitem{robustvision2018}
{Geiger, A. and Nie{\ss}ner, M. and Dai, A.}: Robust {V}ision {C}hallenge
  {CVPR} {W}orkshop (2018)

\bibitem{hao2018detail}
Hao, Z., Li, Y., You, S., Lu, F.: Detail preserving depth estimation from a
  single image using attention guided networks. In: 2018 International
  Conference on 3D Vision (3DV). pp. 304--313. IEEE (2018)

\bibitem{hartley2003multiple}
Hartley, R., Zisserman, A.: Multiple view geometry in computer vision.
  Cambridge university press (2003)

\bibitem{hu2018squeeze}
Hu, J., Shen, L., Sun, G.: Squeeze-and-excitation networks. In: Proceedings of
  the IEEE Conference on Computer Vision and Pattern Recognition. pp.
  7132--7141 (2018)

\bibitem{Hu2018RevisitingSI}
Hu, J., Ozay, M., Zhang, Y., Okatani, T.: Revisiting single image depth
  estimation: Toward higher resolution maps with accurate object boundaries.
  In: IEEE Winter Conf. on Applications of Computer Vision (WACV) (2019)

\bibitem{janoch2013category}
Janoch, A., Karayev, S., Jia, Y., Barron, J.T., Fritz, M., Saenko, K., Darrell,
  T.: A category-level 3d object dataset: Putting the kinect to work. In:
  Consumer Depth Cameras for Computer Vision, pp. 141--165. Springer (2013)

\bibitem{jiao2018look}
Jiao, J., Cao, Y., Song, Y., Lau, R.: Look deeper into depth: Monocular depth
  estimation with semantic booster and attention-driven loss. In: Proceedings
  of the European Conference on Computer Vision (ECCV). pp. 53--69 (2018)

\bibitem{kingma2014adam}
Kingma, D.P., Ba, J.: Adam: A method for stochastic optimization. arXiv
  preprint arXiv:1412.6980  (2014)

\bibitem{Koch18:ECS}
Koch, T., Liebel, L., Fraundorfer, F., K{\"o}rner, M.: Evaluation of cnn-based
  single-image depth estimation methods. In: Leal-Taixé, L., Roth, S. (eds.)
  European Conference on Computer Vision Workshop (ECCV-WS). pp. 331--348.
  Springer International Publishing (2018)

\bibitem{kong2019pixel}
Kong, S., Fowlkes, C.: Pixel-wise attentional gating for scene parsing. In:
  2019 IEEE Winter Conference on Applications of Computer Vision (WACV). pp.
  1024--1033. IEEE (2019)

\bibitem{laina2016deeper}
Laina, I., Rupprecht, C., Belagiannis, V., Tombari, F., Navab, N.: Deeper depth
  prediction with fully convolutional residual networks. In: 2016 Fourth
  International Conference on 3D Vision (3DV). pp. 239--248. IEEE (2016)

\bibitem{lee2019monocular}
Lee, J.H., Kim, C.S.: Monocular depth estimation using relative depth maps. In:
  Proceedings of the IEEE Conference on Computer Vision and Pattern
  Recognition. pp. 9729--9738 (2019)

\bibitem{li2018monocular}
Li, B., Dai, Y., He, M.: Monocular depth estimation with hierarchical fusion of
  dilated {CNNs} and soft-weighted-sum inference. Pattern Recognition
  \textbf{83},  328--339 (2018)

\bibitem{li2018deep}
Li, R., Xian, K., Shen, C., Cao, Z., Lu, H., Hang, L.: Deep attention-based
  classification network for robust depth prediction. In: Asian Conference on
  Computer Vision. pp. 663--678. Springer (2018)

\bibitem{liu2019planercnn}
Liu, C., Kim, K., Gu, J., Furukawa, Y., Kautz, J.: Planercnn: 3d plane
  detection and reconstruction from a single image. In: Proceedings of the IEEE
  Conference on Computer Vision and Pattern Recognition. pp. 4450--4459 (2019)

\bibitem{liu2018planenet}
Liu, C., Yang, J., Ceylan, D., Yumer, E., Furukawa, Y.: Planenet: Piece-wise
  planar reconstruction from a single rgb image. In: Proceedings of the IEEE
  Conference on Computer Vision and Pattern Recognition. pp. 2579--2588 (2018)

\bibitem{NEURIPS2019_9015}
Paszke, A., Gross, S., Massa, F., Lerer, A., Bradbury, J., Chanan, G., Killeen,
  T., Lin, Z., Gimelshein, N., Antiga, L., Desmaison, A., Kopf, A., Yang, E.,
  DeVito, Z., Raison, M., Tejani, A., Chilamkurthy, S., Steiner, B., Fang, L.,
  Bai, J., Chintala, S.: Pytorch: An imperative style, high-performance deep
  learning library. In: Advances in Neural Information Processing Systems 32,
  pp. 8024--8035. Curran Associates, Inc. (2019),
  \url{http://papers.neurips.cc/paper/9015-pytorch-an-imperative-style-high-performance-deep-learning-library.pdf}

\bibitem{qi2018geonet}
Qi, X., Liao, R., Liu, Z., Urtasun, R., Jia, J.: Geonet: Geometric neural
  network for joint depth and surface normal estimation. In: Proceedings of the
  IEEE Conference on Computer Vision and Pattern Recognition. pp. 283--291
  (2018)

\bibitem{ramamonjisoa2019sharpnet}
Ramamonjisoa, M., Lepetit, V.: Sharpnet: Fast and accurate recovery of
  occluding contours in monocular depth estimation. The IEEE International
  Conference on Computer Vision (ICCV) Workshops  (2019)

\bibitem{ren2019deep}
Ren, H., El-khamy, M., Lee, J.: Deep robust single image depth estimation
  neural network using scene understanding. In: Proceedings of the IEEE
  Conference on Computer Vision and Pattern Recognition Workshops. pp. 37--45
  (2019)

\bibitem{saxena2006learning}
Saxena, A., Chung, S.H., Ng, A.Y.: Learning depth from single monocular images.
  In: Advances in neural information processing systems. pp. 1161--1168 (2006)

\bibitem{silberman2012indoor}
Silberman, N., Hoiem, D., Kohli, P., Fergus, R.: Indoor segmentation and
  support inference from rgbd images. In: European Conference on Computer
  Vision. pp. 746--760. Springer (2012)

\bibitem{song2015sun}
Song, S., Lichtenberg, S.P., Xiao, J.: Sun rgb-d: A rgb-d scene understanding
  benchmark suite. In: Proceedings of the IEEE conference on computer vision
  and pattern recognition. pp. 567--576 (2015)

\bibitem{szeliski2011structure}
Szeliski, R.: Structure from motion. In: Computer Vision, pp. 303--334.
  Springer (2011)

\bibitem{tan2019mnasnet}
Tan, M., Chen, B., Pang, R., Vasudevan, V., Sandler, M., Howard, A., Le, Q.V.:
  Mnasnet: Platform-aware neural architecture search for mobile. In:
  Proceedings of the IEEE Conference on Computer Vision and Pattern
  Recognition. pp. 2820--2828 (2019)

\bibitem{wang2017residual}
Wang, F., Jiang, M., Qian, C., Yang, S., Li, C., Zhang, H., Wang, X., Tang, X.:
  Residual attention network for image classification. In: Proceedings of the
  IEEE Conference on Computer Vision and Pattern Recognition. pp. 3156--3164
  (2017)

\bibitem{wang2018non}
Wang, X., Girshick, R., Gupta, A., He, K.: Non-local neural networks. In:
  Proceedings of the IEEE Conference on Computer Vision and Pattern
  Recognition. pp. 7794--7803 (2018)

\bibitem{wofk2019fastdepth}
Wofk, D., Ma, F., Yang, T.J., Karaman, S., Sze, V.: Fastdepth: Fast monocular
  depth estimation on embedded systems. In: 2019 International Conference on
  Robotics and Automation (ICRA). pp. 6101--6108. IEEE (2019)

\bibitem{xiao2013sun3d}
Xiao, J., Owens, A., Torralba, A.: Sun3d: A database of big spaces
  reconstructed using sfm and object labels. In: Proceedings of the IEEE
  International Conference on Computer Vision. pp. 1625--1632 (2013)

\bibitem{xu2018structured}
Xu, D., Wang, W., Tang, H., Liu, H., Sebe, N., Ricci, E.: Structured attention
  guided convolutional neural fields for monocular depth estimation. In:
  Proceedings of the IEEE Conference on Computer Vision and Pattern
  Recognition. pp. 3917--3925 (2018)

\bibitem{xu2015show}
Xu, K., Ba, J., Kiros, R., Cho, K., Courville, A., Salakhudinov, R., Zemel, R.,
  Bengio, Y.: Show, attend and tell: Neural image caption generation with
  visual attention. In: International conference on machine learning. pp.
  2048--2057 (2015)

\bibitem{Yin2019enforcing}
Yin, W., Liu, Y., Shen, C., Yan, Y.: Enforcing geometric constraints of virtual
  normal for depth prediction. In: The IEEE International Conference on
  Computer Vision (ICCV) (2019)

\bibitem{Yu2016}
Yu, F., Koltun, V.: Multi-scale context aggregation by dilated convolutions.
  In: International Conference on Learning Representations (ICLR) (2016)

\bibitem{yu2017dilated}
Yu, F., Koltun, V., Funkhouser, T.: Dilated residual networks. In: Proceedings
  of the IEEE conference on computer vision and pattern recognition. pp.
  472--480 (2017)

\bibitem{zhang2016physically}
Zhang, Y., Song, S., Yumer, E., Savva, M., Lee, J.Y., Jin, H., Funkhouser, T.:
  Physically-based rendering for indoor scene understanding using convolutional
  neural networks. The IEEE Conference on Computer Vision and Pattern
  Recognition (CVPR)  (2017)

\bibitem{zhang2019pattern}
Zhang, Z., Cui, Z., Xu, C., Yan, Y., Sebe, N., Yang, J.: Pattern-affinitive
  propagation across depth, surface normal and semantic segmentation. In:
  Proceedings of the IEEE Conference on Computer Vision and Pattern
  Recognition. pp. 4106--4115 (2019)

\end{thebibliography}

\appendix
\title{- Supplementary Material -\\Guiding Monocular Depth Estimation Using Depth-Attention Volume}
\titlerunning{Supplementary Material}

\author{Lam Huynh\inst{1} \and
Phong Nguyen-Ha\inst{1} \and
Jiri Matas\inst{2} \and
Esa Rahtu\inst{3} \and
Janne Heikkil\"a\inst{1} }
\authorrunning{L. Huynh et al.}

\institute{Center for Machine Vision and Signal Analysis, University of Oulu, Finland \and
Center for Machine Perception, Czech Technical University, Czech Republic \and
Computer Vision Group, Tampere University, Finland }

\maketitle

In this document, we present additional qualitative results on NYU-Depth-v2 and SUN-RGBD datasets in Section~\ref{QualitativeNYU} and~\ref{QualitativeSUN}, respectively. Section~\ref{AdditionalAnalysis} provides extensive analysis for planarity error and non-local embedding space selection strategy. Details of the network architecture are described in Section~\ref{DetailArchitecture} while Section~\ref{DetailMetrics} gives the  definitions of the evaluation metrics. Besides, we attach a video demo of our monocular depth estimation model for a random indoor scene in the supplementary material. \vspace{-0.5em}

\section{Additional qualitative results on NYU-Depth-v2} \label{QualitativeNYU} 

This section provides further results and analysis on NYU-Depth-v2 dataset. \vspace{-1.1em}

\subsection{Depth map with and without $\mathcal{L}_{attention}$}%\vspace{-0.25cm}

\begin{figure}[H]
  \centering
  \scalebox{0.9}{
\begin{tabular}{c c c c}

\includegraphics[width=0.225\textwidth]{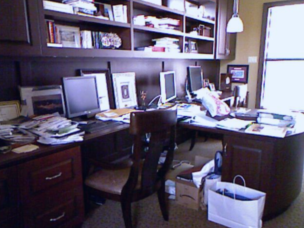}&
\includegraphics[width=0.225\textwidth]{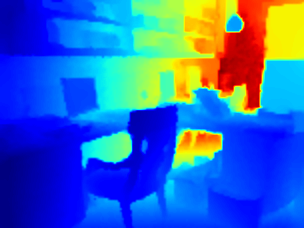}&
\includegraphics[width=0.225\textwidth]{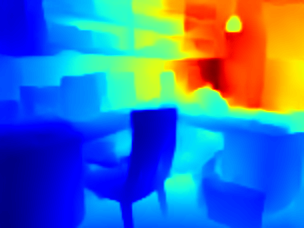}&
\includegraphics[width=0.225\textwidth]{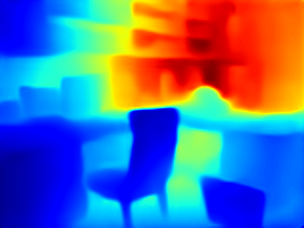} \\

\includegraphics[width=0.225\textwidth]{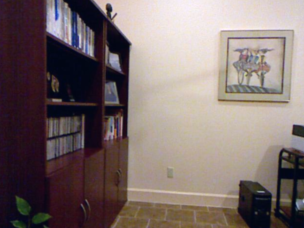}&
\includegraphics[width=0.225\textwidth]{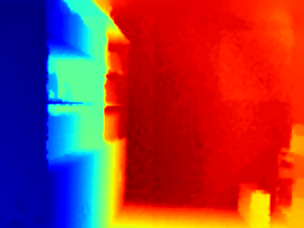}&
\includegraphics[width=0.225\textwidth]{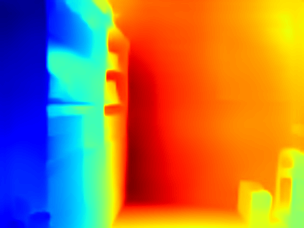}&
\includegraphics[width=0.225\textwidth]{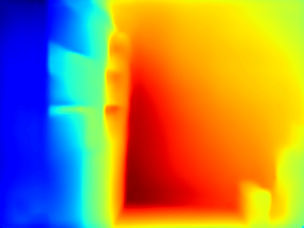} \\

\includegraphics[width=0.225\textwidth]{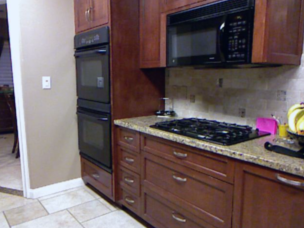}&
\includegraphics[width=0.225\textwidth]{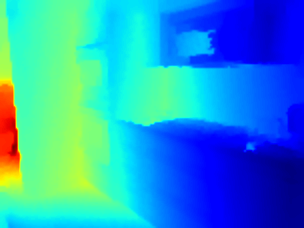}&
\includegraphics[width=0.225\textwidth]{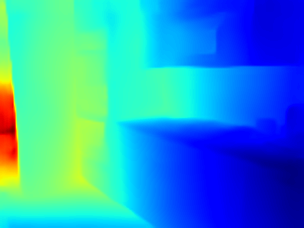}&
\includegraphics[width=0.225\textwidth]{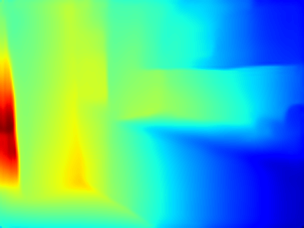} \\

Image & Ground truth & Full loss & w/o $\mathcal{L}_{attention}$ \\

\end{tabular}
}
    \caption{Predicted depth maps from our model train with and without the $\mathcal{L}_{attention}$ term.}%
    \label{fig:more_qualitative_nyu}
\end{figure}

\begin{figure}[H]
  \centering
\begin{tabular}{c c c c}

\includegraphics[width=0.225\textwidth]{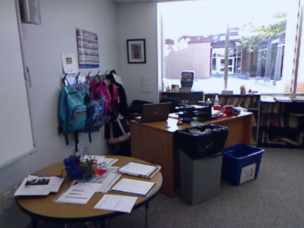}&
\includegraphics[width=0.225\textwidth]{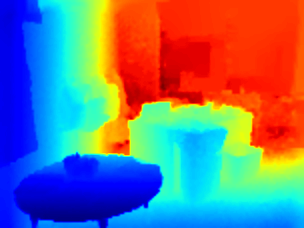}&
\includegraphics[width=0.225\textwidth]{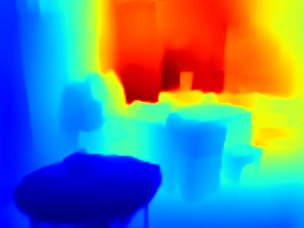}&
\includegraphics[width=0.225\textwidth]{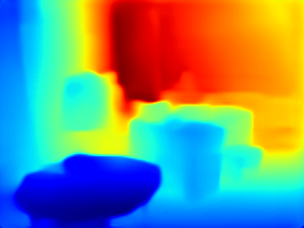} \\

\includegraphics[width=0.225\textwidth]{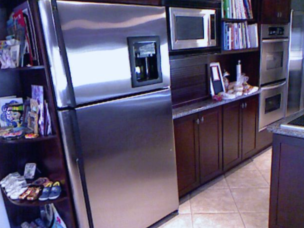}&
\includegraphics[width=0.225\textwidth]{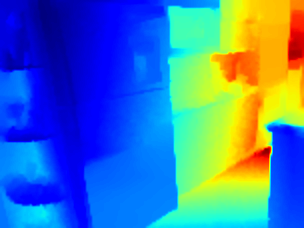}&
\includegraphics[width=0.225\textwidth]{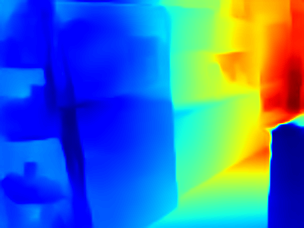}&
\includegraphics[width=0.225\textwidth]{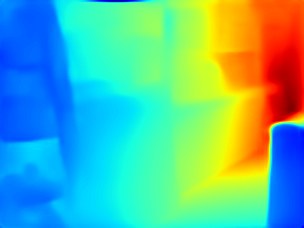} \\

\includegraphics[width=0.225\textwidth]{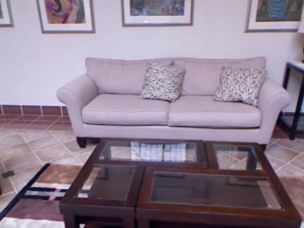}&
\includegraphics[width=0.225\textwidth]{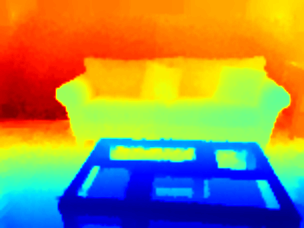}&
\includegraphics[width=0.225\textwidth]{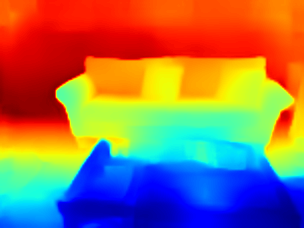}&
\includegraphics[width=0.225\textwidth]{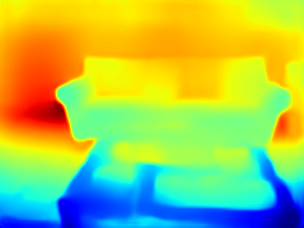} \\

\includegraphics[width=0.225\textwidth]{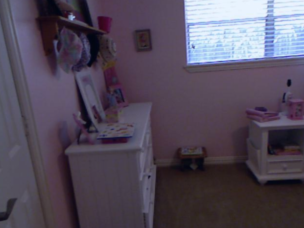}&
\includegraphics[width=0.225\textwidth]{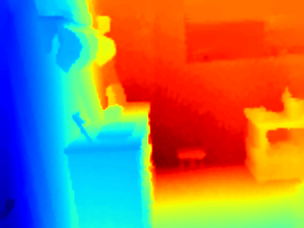}&
\includegraphics[width=0.225\textwidth]{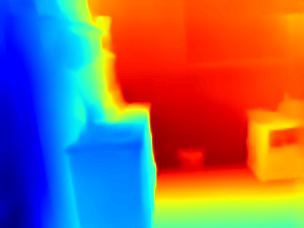}&
\includegraphics[width=0.225\textwidth]{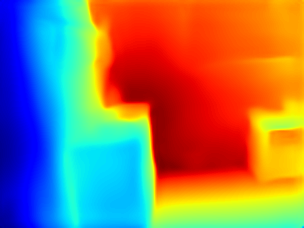} \\

\includegraphics[width=0.225\textwidth]{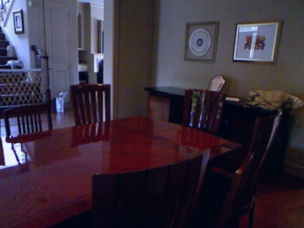}&
\includegraphics[width=0.225\textwidth]{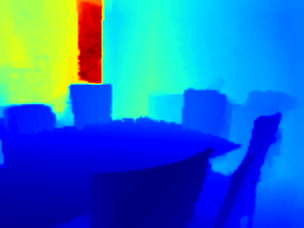}&
\includegraphics[width=0.225\textwidth]{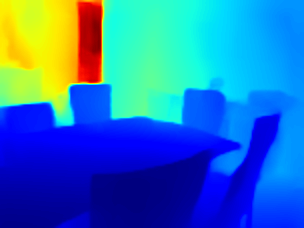}&
\includegraphics[width=0.225\textwidth]{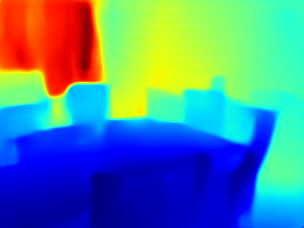} \\

\includegraphics[width=0.225\textwidth]{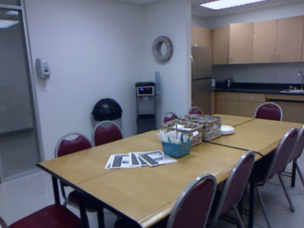}&
\includegraphics[width=0.225\textwidth]{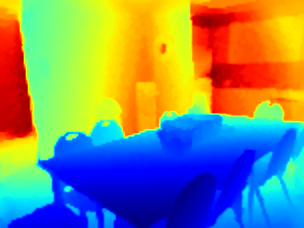}&
\includegraphics[width=0.225\textwidth]{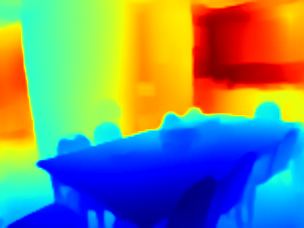}&
\includegraphics[width=0.225\textwidth]{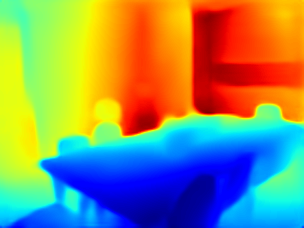} \\

\includegraphics[width=0.225\textwidth]{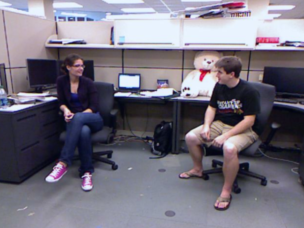}&
\includegraphics[width=0.225\textwidth]{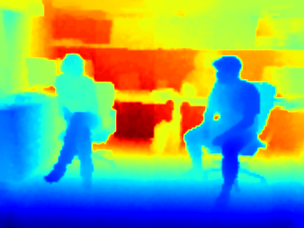}&
\includegraphics[width=0.225\textwidth]{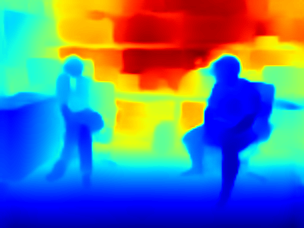}&
\includegraphics[width=0.225\textwidth]{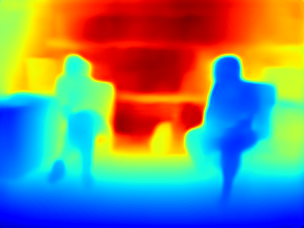} \\

\includegraphics[width=0.225\textwidth]{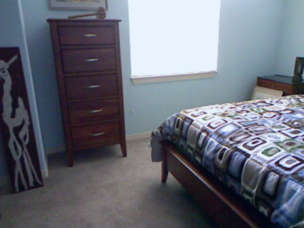}&
\includegraphics[width=0.225\textwidth]{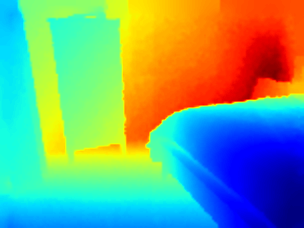}&
\includegraphics[width=0.225\textwidth]{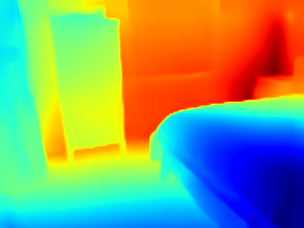}&
\includegraphics[width=0.225\textwidth]{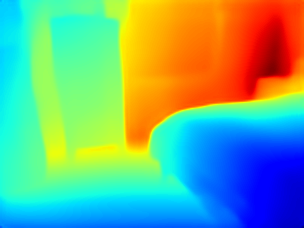} \\

Image & Ground truth & Full loss & w/o $\mathcal{L}_{attention}$ \\

\end{tabular}
    \caption{Predicted depth maps from our model train with and without the $\mathcal{L}_{attention}$ term.}%
    \label{fig:more_qualitative_nyu1}
\end{figure}

As shown in Figure~\ref{fig:more_qualitative_nyu} and~\ref{fig:more_qualitative_nyu1}, the model with full loss significantly improves depth map quality at boundaries and detailed areas.%\vspace{-0.5cm}

\subsection{Point cloud reconstructions}%

\begin{figure}[H]
    \includegraphics[width=0.99\textwidth]{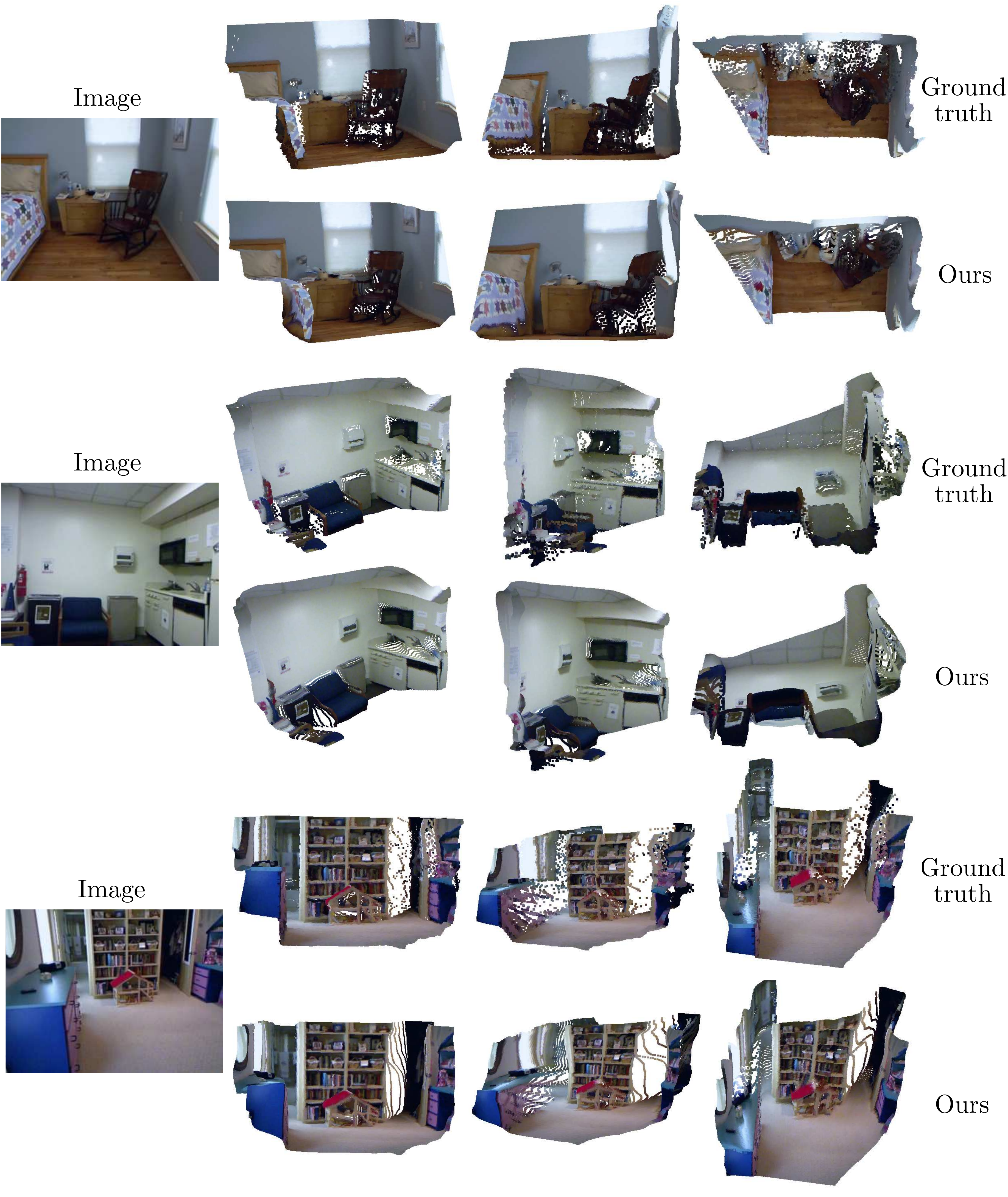}
  \centering
  \caption{Reconstructed point clouds for a set of randomly selected examples from NYU-Depth-v2. The images from the point clouds are captured in different camera poses to provide an overview of the 3D scenes.}%
  \label{fig:nyu_point_cloud}
\end{figure}

We further examine the accuracy of the predicted depth maps by reconstructing the point clouds of three arbitrary views in the NYU-Depth-v2 test set. The back-projected 3D points are shown in Figure~\ref{fig:nyu_point_cloud}. The results near the walls, floors and ceilings are virtually linear and close to the ground truths.

\section{Additional qualitative results of cross-dataset evaluation on SUN-RGBD}\vspace{-0.25cm} \label{QualitativeSUN}

In this section, we use our pretrained model on  NYU-Depth-v2~\citeSupp{silberman2012indoor_supp} to estimate depth values from SUN-RGBD images~\citeSupp{song2015sun_supp,janoch2013category_supp,xiao2013sun3d_supp}. Dissecting the predicted depth maps, reconstruction point clouds and attention maps demonstrates the generalization ability of our proposed method.\vspace{-0.3cm}

\subsection{Predicted depth maps}\vspace{-0.15cm}

\begin{figure}[H]
  \centering
  \scalebox{0.8}{
\begin{tabular}{c c c}

\includegraphics[width=0.225\textwidth]{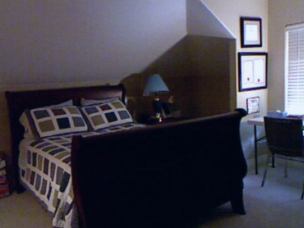}&
\includegraphics[width=0.225\textwidth]{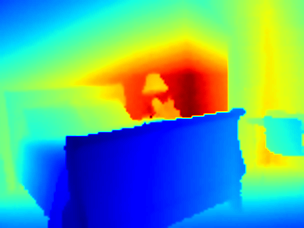}&
\includegraphics[width=0.225\textwidth]{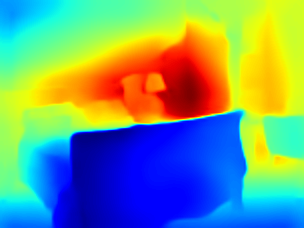} \\

\includegraphics[width=0.225\textwidth]{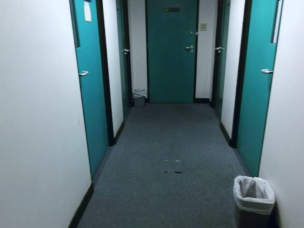}&
\includegraphics[width=0.225\textwidth]{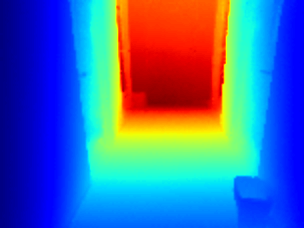}&
\includegraphics[width=0.225\textwidth]{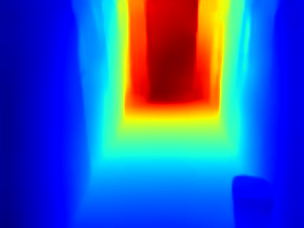} \\

\includegraphics[width=0.225\textwidth]{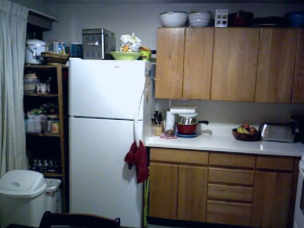}&
\includegraphics[width=0.225\textwidth]{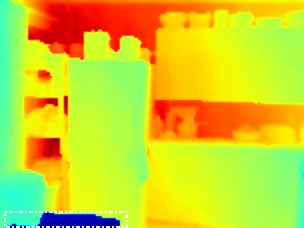}&
\includegraphics[width=0.225\textwidth]{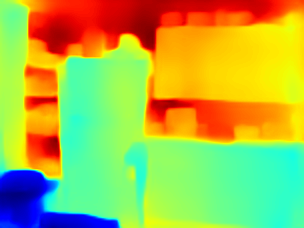} \\

\includegraphics[width=0.225\textwidth]{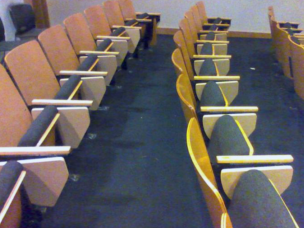}&
\includegraphics[width=0.225\textwidth]{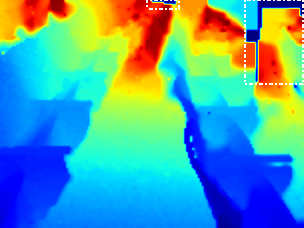}&
\includegraphics[width=0.225\textwidth]{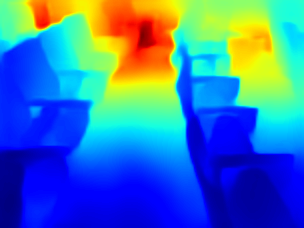} \\

\includegraphics[width=0.225\textwidth]{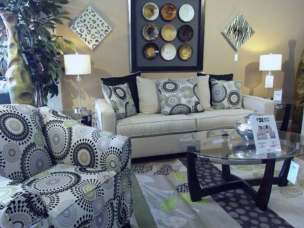}&
\includegraphics[width=0.225\textwidth]{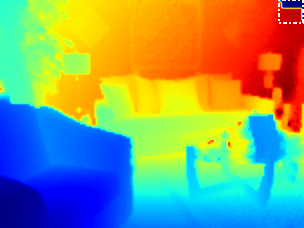}&
\includegraphics[width=0.225\textwidth]{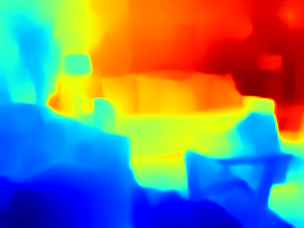} \\

\includegraphics[width=0.225\textwidth]{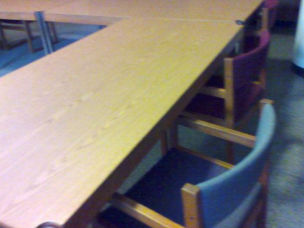}&
\includegraphics[width=0.225\textwidth]{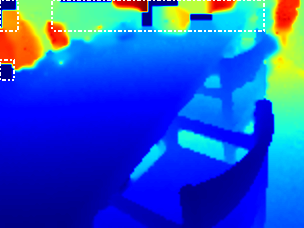}&
\includegraphics[width=0.225\textwidth]{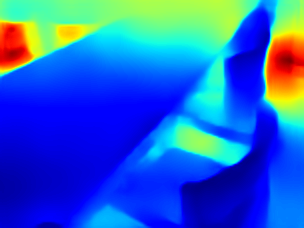} \\

\includegraphics[width=0.225\textwidth]{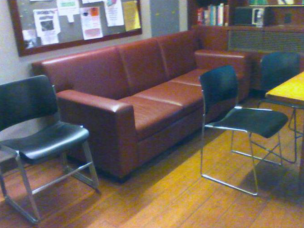}&
\includegraphics[width=0.225\textwidth]{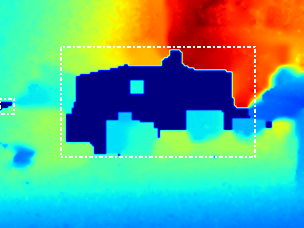}&
\includegraphics[width=0.225\textwidth]{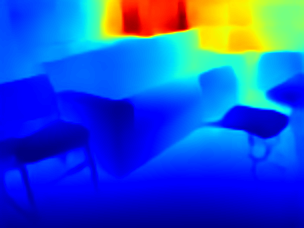} \\

Image & Ground truth & Ours \\

\end{tabular}
}
    \caption{Ramdomly examples from the SUN-RGBD test set. Areas in the white boxes show missing or incorrect depth values from the ground truth data.}
    \label{fig:qualitative_sun_rgbd}
\end{figure}

As shown in Figure~\ref{fig:qualitative_sun_rgbd}, our model provides reasonable depth maps for SUN-RGBD examples although it has not been trained on this dataset. The geometry layout of the scene is retained, even in difficult scenarios (e.g. images in row (5) and (6) in Figure~\ref{fig:qualitative_sun_rgbd}).

\subsection{Analyzing the predicted attention maps}\vspace{-0.1cm}

\begin{figure}[H]
    \includegraphics[width=0.855\textwidth]{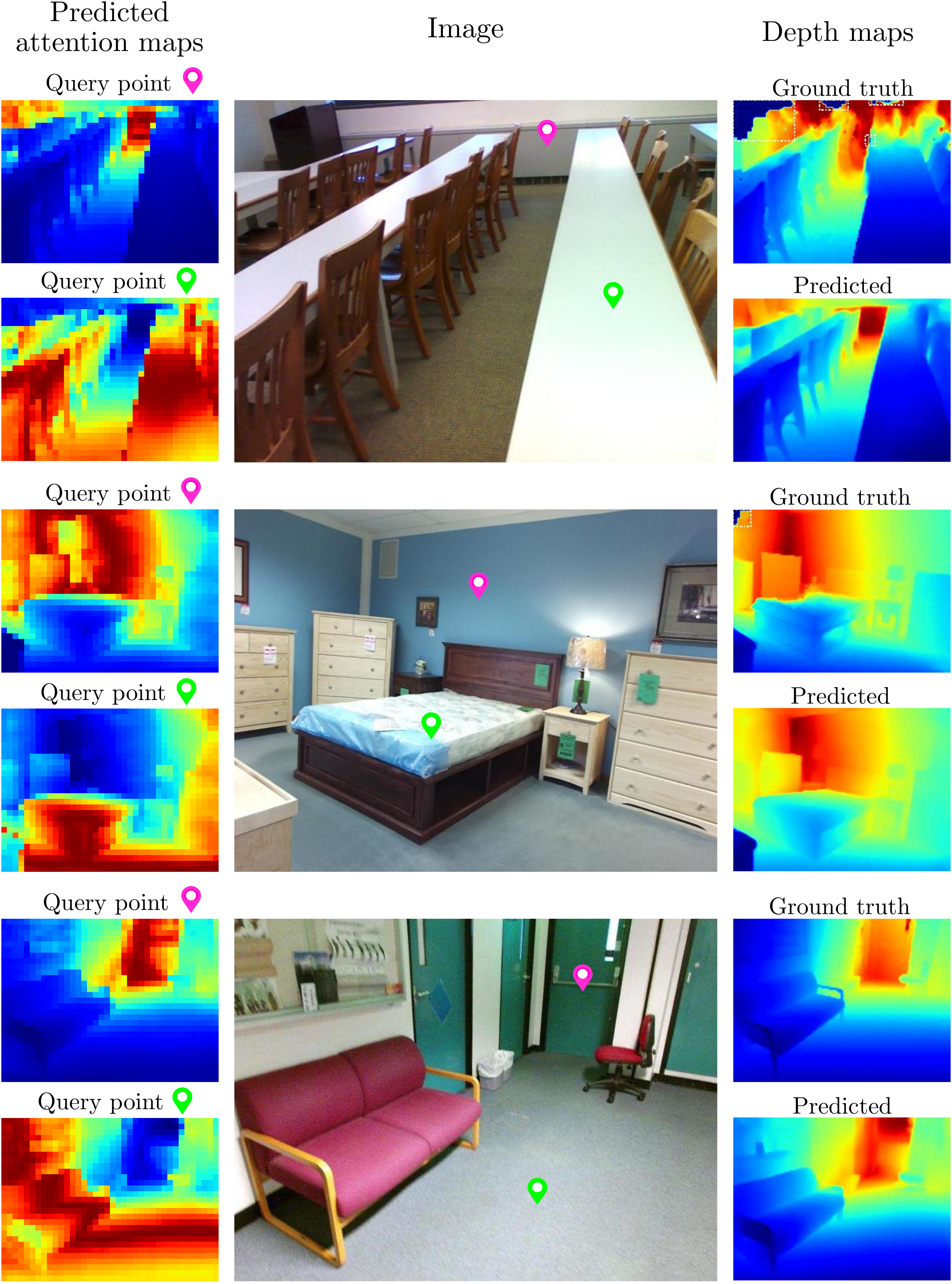}
  \centering
  \caption{Estimated attention and depth maps of our model train on NYU-Depth-v2 and test on SUN-RGBD. Left column presents predicted attention maps at indicated query points, while right column shows the predicted and ground truth depth maps. The input images are displayed in the middle.}%
  \label{fig:sun_attention_maps}
\end{figure}

As depicted in Figure~\ref{fig:sun_attention_maps}, the proposed network learns to pay attention on planar-areas. At the green query point in the first image, the network concentrates on table surfaces as indicated by the warm color in its attention map. At the magenta query point, the model shifts its attention to the wall in the background. %

\subsection{Point cloud reconstruction}

\begin{figure}[H]
    \includegraphics[width=0.9\textwidth]{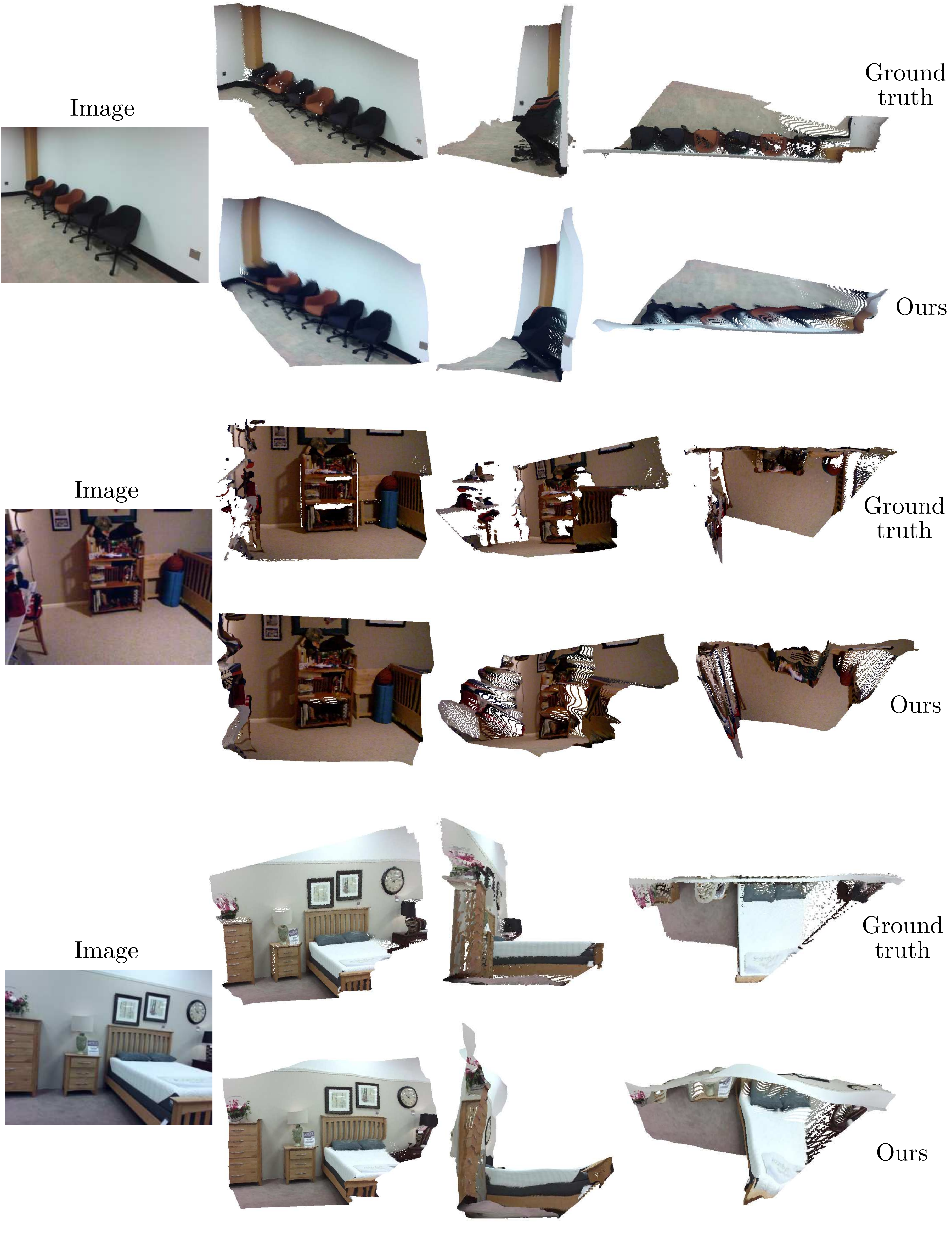}
  \centering
  \caption{Reconstructed point clouds for randomly selected samples from the SUN-RGBD test set. The images of the point clouds are captured from different camera poses to provide an overview of the 3D scenes. The estimated depth maps are obtains from the model train with NYU-Depth-v2. }%
  \label{fig:sun_point_cloud}
\end{figure}

Figure~\ref{fig:sun_point_cloud} illustrates re-projected point clouds for SUN-RGBD examples produced by a model that is trained on NYU-Depth-v2. The produced point clouds are relatively close to the ground truth despite the fact that the model was trained on a different dataset. %

\section{Additional analysis}\vspace{-0.25cm} \label{AdditionalAnalysis}

\subsection{Planarity error}\vspace{-0.15cm}

\begin{table}[t!]
    \caption{iBims-1 benchmark. Metrics with $\downarrow$ mean lower is better and $\uparrow$ mean higher is better. Methods indicated with $^{\dagger}$ and $^{\ddagger}$ are trained using the AlexNet or VGG, respectively.}\vspace{-0.25cm}
    \centering
    \small
    \scalebox{0.83}{%
    \begin{tabular}{|l|c|c|c|c|c|c|c|c|c|c|c|c|c|}
    \hline
    \textbf{Method} & \textbf{REL}$\downarrow$ & \textbf{log10}$\downarrow$ & \textbf{RMS}$\downarrow$ & \(\boldsymbol{\delta_1}\)$\uparrow$ & \(\boldsymbol{\delta_2}\)$\uparrow$ & \(\boldsymbol{\delta_3}\)$\uparrow$ & \(\boldsymbol{\epsilon^{plan}}\)$\downarrow$ & \(\boldsymbol{\epsilon^{orie}}\)$\downarrow$ & \(\boldsymbol{\epsilon^{acc}}\)$\downarrow$ & \(\boldsymbol{\epsilon^{comp}}\)$\downarrow$ & \(\boldsymbol{\epsilon^{0}}\)$\uparrow$ & \(\boldsymbol{\epsilon^{-}}\)$\downarrow$ & \(\boldsymbol{\epsilon^{+}}\)$\downarrow$ \\ \hline
    
    Eigen'14 \citeSupp{eigen2014depth_supp} & 0.32 & 0.17 & 1.55 & 0.36 & 0.65 & 0.84 & 7.70 & 24.91 & 9.97 & 9.99 & 70.37 & 27.42 & 2.22 \\ %
    \hline
    
    Eigen'15 \citeSupp{eigen2015predicting_supp} $^{\dagger}$ & 0.30 & 0.15 & 1.38 & 0.40 & 0.73 & 0.88 & 7.52 & 21.50 & 4.66 & 8.68 & 77.48 & 18.93 & 3.59 \\ %
    \hline
    
    Eigen'15 \citeSupp{eigen2015predicting_supp} $^{\ddagger}$ & 0.25 & 0.13 & 1.26 & 0.47 & 0.78 & 0.93 & \textbf{5.97} & 17.65 & 4.05 & 8.01 & 79.88 & 18.72 & 1.41 \\ %
    \hline
    
    Laina'16 \citeSupp{laina2016deeper_supp} & 0.26 & 0.13 & 1.20 & 0.50 & 0.78 & 0.91 & 6.46 & 19.13 & 6.19 & 9.17 & 81.02 & 17.01 & 1.97 \\ %
    \hline
    
    Liu'15 \citeSupp{liu2015deep_supp} & 0.30 & 0.13 & 1.26 & 0.48 & 0.78 & 0.91 & 8.45 & 28.69 & \textbf{2.42} & \textbf{7.11} & 79.70 & 14.16 & 6.14 \\ %
    \hline
    
    Li'17 \citeSupp{li2017two_supp} & \textbf{0.22} & 0.11 & 1.09 & 0.58 & \textbf{0.85} & \textbf{0.94} & 7.82 & 22.20 & 3.90 & 8.17 & 83.71 & 13.20 & 3.09 \\ %
    \hline
     
    Liu'18 \citeSupp{liu2018planenet_supp} & 0.29 & 0.17 & 1.45 & 0.41 & 0.70 & 0.86 & 7.26 & \textbf{17.24} & 4.84 & 8.86 & 71.24 & 28.36 & \textbf{0.40} \\ %
    \hline
    
    Ramam.'19 \citeSupp{ramamonjisoa2019sharpnet_supp} & 0.26 & 0.11 & 1.07 & \textbf{0.59} & 0.84 & \textbf{0.94} & 9.95 & 25.67 & 3.52 & 7.61 & 84.03 & 9.48 & 6.49 \\ %
    \hline
    
    \textbf{Ours} & 0.24 & \textbf{0.10} & \textbf{1.06} & \textbf{0.59} & 0.84 & \textbf{0.94} & 7.21 & 18.45 & 3.46 & 7.43 & \textbf{84.36} & \textbf{6.84} & 6.27 \\ \hline
    \end{tabular} }
    \label{tab:ibim1_full}
\end{table} 

\begin{figure}[t!]
    \includegraphics[width=0.63\textwidth]{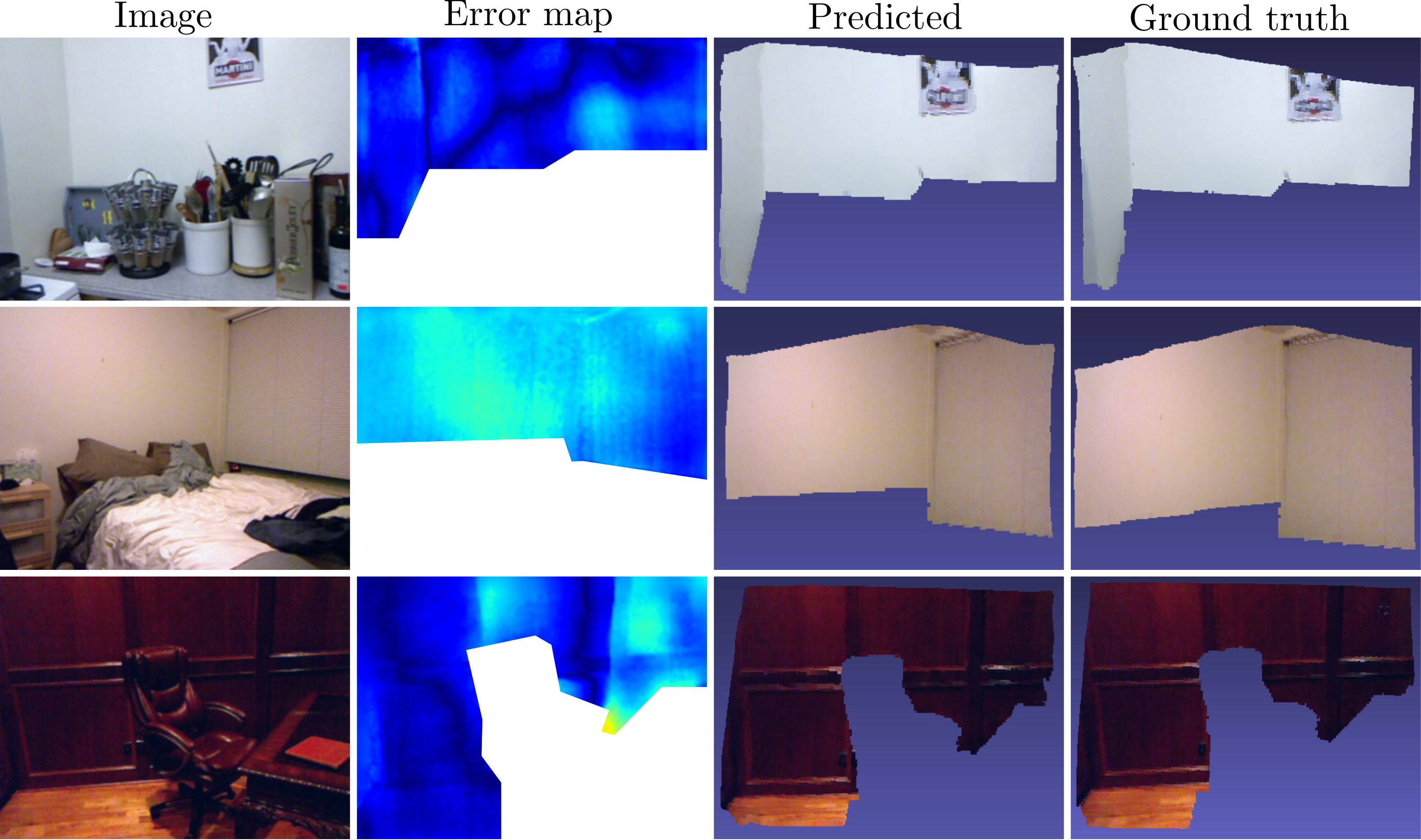}\vspace{-0.3cm}
  \centering
  \caption{Visualization of pixel around the planar areas. The second column shows the error map, while the third and forth column present the predicted and ground truth point cloud.}\vspace{-0.5cm}
  \label{fig:planarity_error}
\end{figure}

Table~\ref{tab:ibim1_full} compares our model with monocular depth estimation methods that officially provides by the iBims-1 benchmark \citeSupp{Koch18:ECS_supp}. The results indicate that we outperform the recent methods \citeSupp{liu2018planenet_supp,ramamonjisoa2019sharpnet_supp} in most of the metrics (including plane related ones). Interestingly, the studies from Li et al. \citeSupp{li2017two_supp} and Liu et al. \citeSupp{liu2015deep_supp} although yield unfavourable results on NYU-Depth-v2 \citeSupp{silberman2012indoor_supp} seem generalize well on the iBims-1. Besides, we show qualitative results of our method around planar areas in Figure~\ref{fig:planarity_error}.

\vspace{-0.25cm}\subsection{Non-local embedding space selection strategy}\vspace{-0.1cm}

\begin{table}[b!]
\caption{\label{tab:cross_modulation}Performance of our model using different types of embedding space.}\vspace{-0.25cm}
\centering
\small
\begin{tabular}{|l|c|c|c|c|c|}
\hline
\textbf{Embedding space} & \textbf{REL$\downarrow$} & \textbf{RMS$\downarrow$} & \(\boldsymbol{\delta_1}\)$\uparrow$ & \(\boldsymbol{\delta_2}\)$\uparrow$ & \(\boldsymbol{\delta_3}\)$\uparrow$ \\ \hline
Single embedding & 0.115 & 0.432 & 0.868 & 0.975 & 0.994 \\ \hline
\textbf{Cross-modulation} & \textbf{0.108} & \textbf{0.412} & \textbf{0.882} & \textbf{0.980} & \textbf{0.996} \\ \hline
\end{tabular}
\end{table}

We empirically found that training the depth attention module using the cross-modulation in two embedding spaces yields superior to using a single embedding with double the number of features as shown in Table~\ref{tab:cross_modulation}.

\section{Network architecture} \label{DetailArchitecture}

This section gives complementary details regarding the network architecture and training process. The general structure of our network encompasses an encoder, a non-local depth attention module and a decoder. We construct the encoder by removing the average pooling and fully connected layer from the DRN-D-22 variation of the dilated residual networks \citeSupp{Yu2016_supp,yu2017dilated_supp}. Table~\ref{tab:encoder_detail} shows the detailed structure of our encoder where \textbf{\textit{conv}} represents 2D convolutional layer with specific kernel-size (\textbf{\textit{k}}), stride (\textbf{\textit{s}}), and dilation (\textbf{\textit{d}}). \textbf{\textit{bn}} stands for batch normalization. \textbf{CH} is the number of output channels and \textbf{\textit{RES}} is the spatial resolution of the output feature maps. \textbf{\textit{basic-block}} represents the basic residual block with corresponding dilation.

As explained in the manuscript, we split the training scheme into three parts. It is worth to mention that during the first training phase, we initialize the encoder with pre-trained weights on the ImageNet \citeSupp{deng2009imagenet_supp}. Our experiments confirm that using the pretraining model improves accuracy and speed of convergence. The second and third training stages follow the procedure described in the main paper.

\begin{table}[H]
\caption{\label{tab:encoder_detail}Detail structure of the encoder. }
\begin{center}
\begin{tabular}{ m{2.8cm} >{\centering\arraybackslash}m{2.15cm} >{\centering\arraybackslash}m{0.5cm} >{\centering\arraybackslash}m{0.4cm} >{\centering\arraybackslash}m{0.4cm} >{\centering\arraybackslash}m{0.4cm} >{\centering\arraybackslash}m{2.0cm} m{2.0cm} }
\multicolumn{8}{c}{\textit{Encoder}} \\ \hline
\textbf{Input} & \textbf{Operations} & \textbf{\textit{k}} & \textbf{\textit{s}} & \textbf{\textit{d}} & \textbf{CH} & \textbf{RES} & \textbf{Output} \\ \hline
image & conv+bn+relu & 7 & 1 & 1 & 16 & $228 \times 304$ & layer0 \\
layer0 & conv+bn+relu & 3 & 1 & 1 & 16 & $228 \times 304$ & layer1 \\
layer1 & conv+bn+relu & 3 & 2 & 1 & 32 & $114 \times 152$ & layer2 \\ \hline
layer2 & basic-block & - & - & 1 & 64 & $57 \times 76$ & d-res-1a \\
d-res-1a & conv+bn & 1 & 2 & 1 & 64 & $57 \times 76$ & d-res-1b \\
d-res-1b & basic-block & - & - & 1 & 64 & $57 \times 76$ & layer3 \\ \hline
layer3 & basic-block & - & - & 1 & 128 & $29 \times 38$ & d-res-2a \\
d-res-2a & conv+bn & 1 & 2 & 1 & 128 & $29 \times 38$ & d-res-2b \\
d-res-2b & basic-block & - & - & 1 & 128 & $29 \times 38$ & layer4 \\ \hline
layer4 & basic-block & - & - & 2 & 256 & $29 \times 38$ & d-res-3a \\
d-res-3a & conv+bn & 1 & 1 & 1 & 256 & $29 \times 38$ & d-res-3b \\
d-res-3b & basic-block & - & - & 2 & 256 & $29 \times 38$ & layer5 \\ \hline
layer5 & basic-block & - & - & 4 & 512 & $29 \times 38$ & d-res-4a \\
d-res-4a & conv+bn & 1 & 1 & 1 & 512 & $29 \times 38$ & d-res-4b \\
d-res-4b & basic-block & - & - & 4 & 512 & $29 \times 38$ & layer6 \\ \hline
layer6 & conv+bn+relu & 3 & 1 & 2 & 512 & $29 \times 38$ & layer7 \\
layer7 & conv+bn+relu & 3 & 1 & 1 & 512 & $29 \times 38$ & layer8-X \\ \hline

\end{tabular}
\end{center}
\end{table}

The non-local depth attention module is the central component of our network with the detailed structure provided in Table~\ref{tab:non_local_detail}. In that, \textbf{green, blue, orange} indicate the green, blue, and orange embedding spaces mentioned in the manuscript. ``$\bigodot$" denotes element-wise multiplication, ``$\bigoplus$" indicates element-wise sum, and ``$\bigotimes$" is the outer product. Layers denoted with $^{\ddagger\ddagger}$ imply reshaping and permuting the tensor to match the required shape for operation. Note that \textit{green-1bn, green-1$\gamma$, green-1$\beta$} and \textit{blue-1bn, blue-1$\gamma$, blue-1$\beta$} are generated at the same time as indicated by the dashed line.  

\begin{table}[H]
\caption{\label{tab:non_local_detail}Internal structure of the non-local depth attention module. }
\centering
\begin{tabular}{ m{2.95cm} >{\centering\arraybackslash}m{2.0cm} >{\centering\arraybackslash}m{0.5cm} >{\centering\arraybackslash}m{0.45cm} >{\centering\arraybackslash}m{0.45cm} >{\centering\arraybackslash}m{0.45cm} >{\centering\arraybackslash}m{2.0cm} m{2.25cm} }
\multicolumn{8}{c}{\textit{Non-local depth attention module}} \\ \hline
\textbf{Input} & \textbf{Operations} & \textbf{\textit{k}} & \textbf{\textit{s}} & \textbf{\textit{d}} & \textbf{CH} & \textbf{RES} & \textbf{Output} \\ \hline

layer8-X & conv & 1 & 1 & 1 & 256 & $29 \times 38$ & orange \\ \hline
layer8-X & conv & 1 & 1 & 1 & 1024 & $29 \times 38$ & green-1 \\
green-1 & bn & - & - & - & 1024 & $29 \times 38$ & green-1bn \\
green-1 & conv & 1 & 1 & 1 & 1024 & $29 \times 38$ & green-1$\gamma$ \\
green-1 & conv & 1 & 1 & 1 & 1024 & $29 \times 38$ & green-1$\beta$ \\ \hdashline
green-1bn, blue-1$\gamma$ & $\bigodot$ & - & - & - & 1024 & $29 \times 38$ & green-1bn-$\gamma$ \\
green-1bn-$\gamma$, blue-1$\beta$ & $\bigoplus$ & - & - & - & 1024 & $29 \times 38$ & green-1-denorm \\
green-1-denorm & relu+conv & 1 & 1 & 1 & 1024 & $29 \times 38$ & green-2 \\ \hline
layer8-X & conv & 1 & 1 & 1 & 1024 & $29 \times 38$ & blue-1 \\ 
blue-1 & bn & - & - & - & 1024 & $29 \times 38$ & blue-1bn \\ 
blue-1 & conv & 1 & 1 & 1 & 1024 & $29 \times 38$ & blue-1$\gamma$ \\ 
blue-1 & conv & 1 & 1 & 1 & 1024 & $29 \times 38$ & blue-1$\beta$ \\ \hdashline
blue-1bn, green-1$\gamma$ & $\bigodot$ & - & - & - & 1024 & $29 \times 38$ & blue-1bn-$\gamma$ \\
blue-1bn-$\gamma$, green-1$\beta$ & $\bigoplus$ & - & - & - & 1024 & $29 \times 38$ & blue-1-denorm \\
blue-1-denorm & relu+conv & 1 & 1 & 1 & 1024 & $29 \times 38$ & blue-2 \\ \hline
green-2$^{\ddagger\ddagger}$, blue-2$^{\ddagger\ddagger}$ & $\bigotimes$ & - & - & - & 1 & $1102 \times 1102$ & dav-1 \\
dav-1 & sigmoid & - & - & - & 1 & $1102 \times 1102$ & dav-2 \\ 
dav-2, orange & $\bigotimes$ & - & - & - & 256 & $29 \times 38$ & dav-3$^{\ddagger\ddagger}$ \\
dav-3 & conv+bn & 1 & 1 & 1 & 512 & $29 \times 38$ & dav-4 \\
dav-4, layer8-X & $\bigoplus$ & - & - & - & 512 & $29 \times 38$ & layer8-Y \\ \hline

\end{tabular}
\end{table}

Unlike previous studies \citeSupp{eigen2015predicting_supp,laina2016deeper_supp,hao2018detail_supp,ren2019deep_supp,zhang2019pattern_supp,ramamonjisoa2019sharpnet_supp,Hu2018RevisitingSI_supp,chen2019structure_supp,Yin2019enforcing_supp}, we implement a straightforward decoder with two bilinear upsamplings follow by 2D convolutional layers and batch-normalizations. Finally, the upsampled feature maps are refined to produce the final depth map using two 2D convolutional layers. Table~\ref{tab:decoder_detail} provides a detailed structure of our decoder. 

\begin{table}[H]
\caption{\label{tab:decoder_detail}Internal structure of the decoder where \textbf{bilinear} represents bilinear upsampling layers.}
\centering
\begin{tabular}{ m{2.4cm} >{\centering\arraybackslash}m{2.6cm} >{\centering\arraybackslash}m{0.5cm} >{\centering\arraybackslash}m{0.4cm} >{\centering\arraybackslash}m{0.4cm} >{\centering\arraybackslash}m{0.4cm} >{\centering\arraybackslash}m{2.0cm} m{2.0cm} }
\multicolumn{8}{c}{\textit{Decoder}} \\ \hline
\textbf{Input} & \textbf{Operations} & \textbf{\textit{k}} & \textbf{\textit{s}} & \textbf{\textit{d}} & \textbf{CH} & \textbf{RES} & \textbf{Output} \\ \hline

layer8-Y & bilinear+conv+bn & 3 & 1 & 1 & 256 & $57 \times 76$ & up-conv-1 \\
up-conv-1 & bilinear+conv+bn & 3 & 1 & 1 & 128 & $114 \times 152$ & up-conv-2 \\ \hline
up-conv-2 & conv+bn+relu & 5 & 1 & 1 & 64 & $114 \times 152$ & refine-1 \\
refine-1 & conv & 5 & 1 & 1 & 1 & $114 \times 152$ & depth \\ \hline

\end{tabular}
\end{table}

\section{Definitions of the evaluation metrics} \label{DetailMetrics}
All pixels in the predicted and ground truth depth maps with depth values in the range [0.0, 10.0]  are considered valid and used to calculate the errors. We evaluate the performance for our model and for baselines on the NYU-Depth-v2 \citeSupp{silberman2012indoor_supp} using the following metrics:

\begin{itemize}
    \item Mean absolute relative error (REL):
    \begin{equation} 
        \label{eq:metric_rel}
        \frac{1}{T} \sum_{i=1}^{T}{ \frac{|\hat{d}_i - d|}{d} }
    \end{equation}
    
    \item Root mean square error (RMSE):
    \begin{equation} 
        \label{eq:metric_rmse}
        \sqrt[]{\frac{1}{T} \sum_{i=1}^{T}{(\hat{d}_i - d_i)^2 }}
    \end{equation}
    
    \item Thresholded accuracy ($\delta_i$):
    \begin{equation} 
        \label{eq:metric_thres_acc}
        max(\frac{\hat{d}_i}{d_i}, \frac{d_i}{\hat{d}_i}) = \delta^{i} < 1.25^i \quad (i=1,2,3)
    \end{equation}
\end{itemize}

\noindent where $T$ is the number of valid pixel, \(\hat{d}_i\) indicates the predicted depth value at pixel $i$, and \(d_i\) is the ground truth depth at pixel $i$. Lower REL and RMSE values indicate better results, while the higher $\delta_1$, $\delta_2$ and $\delta_3$, the better. In addition to the mean absolute relative error (REL), we assess model performance on ScanNet \citeSupp{dai2017scannet_supp} and SUN-RGBD \citeSupp{song2015sun_supp,janoch2013category_supp,xiao2013sun3d_supp} using:

\begin{itemize}
    \item Mean relative square error (sqREL):
    \begin{equation} 
        \label{eq:metric_sq_rel}
        \frac{1}{T} \sum_{i=1}^{T}{ \frac{ (\hat{d}_i - d_i)^2}{d_i^2} }
    \end{equation}
    
    \item Mean absolute error of the inverse depth (iMAE):
    \begin{equation} 
        \label{eq:metric_inv_mae}
        \frac{1}{T} \sum_{i=1}^{T}{|\hat{p}_i - p_i| }
    \end{equation}
    
    \item Root mean square error of the inverse depth (iRMSE):
    \begin{equation} 
        \label{eq:metric_inv_rmse}
        \sqrt[]{\frac{1}{T} \sum_{i=1}^{T}{(\hat{p}_i - p_i)^2 }}
    \end{equation}
    
    \item Scale-invariant mean square error (SI)~\citeSupp{eigen2014depth_supp}:
    \begin{equation} 
        \label{eq:metric_si}
        \frac{1}{2T} \sum_{i=1}^{T}{ \left[ \log{\hat{d_i}} - \log{d_i} + \frac{1}{T}\sum_{j=1}^{T}{(\log{d_j} - \log{\hat{d_j}})} \right] ^2}
    \end{equation}
\end{itemize}

\noindent where \(\hat{p}_i\) and \(p_i\) are the inverse value of the predicted and ground truth depth at pixel $i$, respectively. For the iBims-1 benchmark \citeSupp{Koch18:ECS_supp}, besides the mentioned metrics, we evaluate model performance using:

\begin{itemize}
    \item Root mean square error in logarithm space (log10):
    \begin{equation} 
        \label{eq:metric_log10}
        \sqrt[]{\frac{1}{T} \sum_{i=1}^{T}{(\log\hat{d}_i - \log d_i)^2 }}
    \end{equation}
    
    \item Flatness of the predicted 3D planes, which measures by the standard deviation of average distance between the predicted 3D points with its corresponding 3D plane ($\epsilon^{plan}$):
    \begin{equation} 
        \label{eq:plane_flatness_error}
        \mathbb{V} \Bigg[\sum_{\textbf{P}_{k;i,j} \in P_k} d(\pi_k, \textbf{P}_{k;i,j}) \Bigg]
    \end{equation}
    
    \item Orientation of the predicted 3D planes, which measures by angle between predicted and ground truth normal vectors ($\epsilon^{orie}$):
    \begin{equation} 
        \label{eq:plane_orie_error}
        acos (n_k, \hat{n}_k)
    \end{equation}
    
\noindent where $\pi_k=(n_k, o_k)$ is the predicted plane with normal vector $n$ and offset $o$, $\textbf{P}_{k;i,j}$ is the 3D point with respect to plane $k^{th}$, and $P_k$ indicates the annotated planes. 
    
    \item Accuracy of depth boundary, which measures by multiplying the predicted edge map with a pre-defined distance map. ($\epsilon^{acc}$):
    \begin{equation} 
        \label{eq:depth_boundary_acc}
        \frac{1}{\sum_i \sum_j \hat{y}_{i,j}} \sum_i \sum_j e_{i,j} \cdot \hat{y}_{i,j}
    \end{equation}
    
    \item Completeness of depth boundary, which measures by multiplying the ground truth edge map with a predicted distance map. ($\epsilon^{comp}$):
    \begin{equation} 
        \label{eq:depth_boundary_comp}
        \frac{1}{\sum_i \sum_j y_{i,j}} \sum_i \sum_j \hat{e}_{i,j} \cdot y_{i,j}
    \end{equation}
    
\noindent where $y$ and $\hat{y}$ are the predicted and ground truth binary edge maps. $e$ and $\hat{e}$ are the pre-defined distance maps which are calculated using the Euclidean distance transform.

\item Directed depth errors ($\epsilon^{0}, \epsilon^{-}, \epsilon^{+}$) are measured based on a reference plane that located at 3 meters distance. The $\epsilon^{0}$ is the percentage of predicted 3D points that lie in the reference plane. On the other hand, $\epsilon^{-}$ and $\epsilon^{+}$ are the propositions of 3D points that lie in front or behind the reference plane.
    
\end{itemize}

\clearpage

\bibliographystyleSupp{splncs04}
\bibliographySupp{ms}

\end{document}